\documentclass[10pt,journal,compsoc]{IEEEtran}
\ifCLASSOPTIONcompsoc
  \usepackage[nocompress]{cite}
\else
  \usepackage{cite}
\fi
\ifCLASSINFOpdf
\else
\fi
\usepackage{graphicx}
\usepackage{xcolor}
\usepackage{color}
\usepackage{mathrsfs}
\usepackage{amsmath}
\usepackage{amssymb}
\usepackage{ragged2e}
\usepackage{multirow}
\usepackage{multicol}
\usepackage{amssymb}
\usepackage{listings}
\usepackage{subcaption}
\usepackage{bbm}
\usepackage[breaklinks=true,bookmarks=true,citecolor=citecolor,colorlinks,pagebackref=true]{hyperref}
\usepackage{colortbl}
\usepackage{float}
\usepackage{dsfont}
\def\ie{\emph{i.e.}}
\def\eg{\emph{e.g.}}
\def\etal{{\em et al.}}
\def\etc{{\em etc.}} 
\definecolor{dkgreen}{rgb}{0,0.6,0}
\definecolor{gray}{rgb}{0.55,0.55,0.55}
\definecolor{mauve}{rgb}{0.58,0,0.82}
\definecolor{citecolor}{HTML}{0071bc}
\definecolor{shadecolor}{rgb}{0.94,0.94,0.94}

\usepackage{wrapfig}
\usepackage[capitalize]{cleveref}  
\crefname{section}{Sec.}{Secs.}
\Crefname{section}{Section}{Sections}
\crefname{table}{Tab.}{Tabs.}
\Crefname{table}{Table}{Tables}
\crefname{figure}{Fig.}{Figs.}
\Crefname{figure}{Figure}{Figures}
\crefname{equation}{Eq.}{Eqs.}
\Crefname{equation}{Equation}{Equations}
\usepackage{diagbox}

\newcolumntype{g}{>{\columncolor{shadecolor}}c}
\usepackage{algorithmic}
\usepackage{algorithm}
\usepackage{booktabs}
\usepackage{array, caption, threeparttable}
\usepackage[font=small,labelfont=bf]{caption}
\newcommand{\method}{FocalClick-XL~}
\begin{document}
\title{FocalClick-XL: Towards Unified and High-quality
Interactive Segmentation}

\def\Eg{\emph{E.g}.}
\def\ie{\emph{i.e}.}
\def\eg{\emph{e.g}.}
\def\ie{\emph{i.e}.}
\def\etal{\emph{et al.}}
\def\cf{\emph{c.f}.}
\def\etc{\emph{etc}}

\author{
Xi Chen, Hengshuang Zhao
\IEEEcompsocitemizethanks{

\IEEEcompsocthanksitem X. Chen and H. Zhao are with the Department of Computer Science, The University of Hong Kong. Email: chauncey0620@gmail.com, hszhao@cs.hku.hk.

\IEEEcompsocthanksitem Corresponding author: Hengshuang Zhao.
}
}

\markboth{}
{Shell \MakeLowercase{\textit{et al.}}: Bare Demo of IEEEtran.cls for Computer Society Journals}

\IEEEtitleabstractindextext{
\begin{abstract}
\justifying
Interactive segmentation enables users to extract binary masks of target objects through simple interactions such as clicks, scribbles, and boxes. However, existing methods often support only limited interaction forms and struggle to capture fine details. In this paper, we revisit the classical coarse-to-fine design of FocalClick and introduce significant extensions. Inspired by its multi-stage strategy, we propose a novel pipeline, \method, to address these challenges simultaneously.
Following the emerging trend of large-scale pretraining, we decompose interactive segmentation into meta-tasks that capture different levels of information—context, object, and detail—assigning a dedicated subnet to each level. This decomposition allows each subnet to undergo scaled pretraining with independent data and supervision, maximizing its effectiveness. To enhance flexibility, we share context- and detail-level information across different interaction forms as common knowledge while introducing a prompting layer at the object level to encode specific interaction types.
As a result, \method achieves state-of-the-art performance on click-based benchmarks and demonstrates remarkable adaptability to diverse interaction formats, including boxes, scribbles, and coarse masks. Beyond binary mask generation, it is also capable of predicting alpha mattes with fine-grained details, making it a versatile and powerful tool for interactive segmentation.

\end{abstract}

}

\maketitle
\IEEEdisplaynontitleabstractindextext
\IEEEpeerreviewmaketitle

\IEEEtitleabstractindextext{
\begin{abstract}
\justifying
Interactive segmentation enables users to extract binary masks of target objects through simple interactions such as clicks, scribbles, and boxes. However, existing methods often support only limited interaction forms and struggle to capture fine details. In this paper, we revisit the classical coarse-to-fine design of FocalClick and introduce significant extensions. Inspired by its multi-stage strategy, we propose a novel pipeline, \method, to address these challenges simultaneously.
Following the emerging trend of large-scale pretraining, we decompose interactive segmentation into meta-tasks that capture different levels of information—context, object, and detail—assigning a dedicated subnet to each level. This decomposition allows each subnet to undergo scaled pretraining with independent data and supervision, maximizing its effectiveness. To enhance flexibility, we share context- and detail-level information across different interaction forms as common knowledge while introducing a prompting layer at the object level to encode specific interaction types.
As a result, \method achieves state-of-the-art performance on click-based benchmarks and demonstrates remarkable adaptability to diverse interaction formats, including boxes, scribbles, and coarse masks. Beyond binary mask generation, it is also capable of predicting alpha mattes with fine-grained details, making it a versatile and powerful tool for interactive segmentation.

\end{abstract}

}
\section{Introduction}
\IEEEPARstart{I}nteractive image segmentation requires users to indicate the target by providing simple indications such as boxes~\cite{iog, sam}, scribbles~\cite{bai2014error,appearancesimilarity,DeepIGeoS,IFIS}, and clicks~\cite{sofiiuk2021ritm, liu2023simpleclick, focalclick, chen2021cdnet}. Compared with traditional annotation tools like the lasso or brush, interactive models could largely reduce the cost of creating masks, which is important for efficient annotation in the era of big data.  

Initial works for interactive segmentation~\cite{liu2023simpleclick,focalclick, sofiiuk2021ritm} mainly focus on designing delicate model structures or developing better strategies~\cite{mahadevan2018iteratively, firstclick} for simulating user interactions. 
However, training on limited data constrains the power of those solutions as the segmentation target could be ``anything''.
Recently, Segment Anything Model~(SAM)~\cite{sam} tackles interactive segmentation with a data-centric solution. It iteratively collects masks with humans in the loop, and finally collects billions of high-quality samples and trains a powerful segmentation model. The generalized strong priors make SAM a foundation solution for interactive segmentation and downstream applications.

However, SAM is not a perfect solution for interactive segmentation as it makes some compromises for ``automatic segmentation''.
First, SAM treats each part of the full image equally without any design of object-centric modeling or detail refinements, which are proven crucial for the fine details in previous interactive segmentation works~\cite{focalclick,hao2021edgeflow,firstclick,fbrs}. The compromise is that SAM considers the computation burden for each independent interaction as it requires densely sampled points for automatic segmentation. Thus, SAM develops a huge encoder to extract the common features for each part of the image and leaves a small prompt encoder to model each interaction. Besides, SAM only supports click and box, and it is hard for the current prompt encoder to encode more flexible user interactions like scribbles and coarse masks\footnote{SAM supports the previous/initial masks as inputs, but shows very limited improvements w/o additional click or box.}.
Different from SAM, we do not consider automatic segmentation but focus on building a state-of-the-art interactive segmentation system.

\begin{figure}[t]
\centering 
\includegraphics[width=1.0\linewidth]{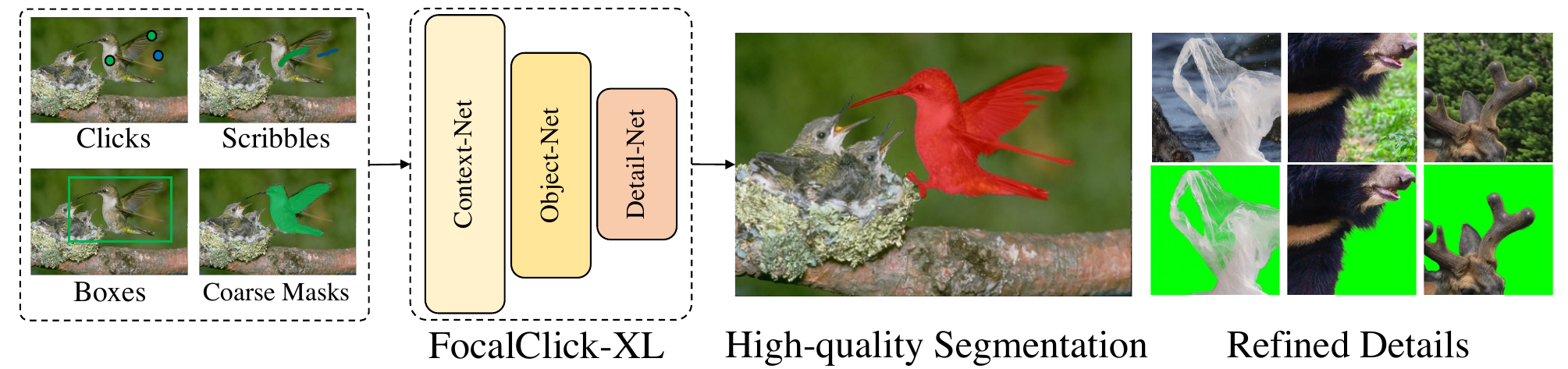} 
\vspace{-8pt}
\caption{%
    \textbf{Demonstrations for \method.} 
     Our method is compatible with various formats of user interactions like clicks, scribbles, boxes, coarse masks, \textit{etc.}, and could predict highly-refined details for both transparent and solid objects.
}
\vspace{-10pt}
\label{fig:teaser}
\end{figure}

Among previous interactive segmentation methods, FocalClick~\cite{focalclick} designs an efficient pipeline following the coarse-to-fine fashion. In this work, we make extensions based on this classical framework and propose \method for high-quality and unified interactive segmentation. Our main insight is to decompose the interactive segmentation pipeline into different subnets to focus on context, object, and detail-level information. This modular design brings the following advantages.

First, each subnet could be pre-trained sufficiently with large-scale specific data, which assists our \method for high-quality segmentation.
Specifically, we leverage the pre-trained visual encoder of SAM for the Context-Net, which takes the full image as input for capturing the global context.  
The Object-Net takes the image patch around the segmentation target and produces the primitive segmentation masks according to the provided user interactions. This part is trained on a combination dataset with object-centric images.
The Detail-Net takes charge of refining the predicted details. It takes the small image patches around the boundaries or low-confident regions as input and is trained on a combined dataset of image segmentation and matting samples.   

Second, in this decomposed pipeline, only the Object-Net is interaction-sensitive, thus we could share the majority of knowledge~(\textit{i.e.}, Context-Net and Detail-Net) across different interactions. It significantly eases the goal of building a unified framework for various interaction forms. Concretely, we add a single Prompting Layer at the input of the Object-Net to encode different types of interactions. To support novel interaction forms, we only need to train a novel Prompting Layer and keep other parameters frozen.

To mitigate the computation burden brought by the cascaded subnet, we follow the strategy of FocalClick. During inference, the Context-Net runs only once per image.  After each round of user interaction, we dynamically zoom in on the object patch and detail patch with a small input size to accelerate the Object-Net and Detail-Net. 
Besides the exploration of the model structure, we thoroughly investigate how to evaluate each type of interaction and propose a series of evaluation protocols and benchmarks. For example, we developed a deterministic scribble generator that supports evaluating scribble-based methods automatically. Besides, we construct a series of coarse mask sets with different qualities to evaluate the robustness for the task of coarse mask refinement.

As presented in \cref{fig:teaser}. \method is compatible with various of user interactions and could predict high-quality masks with fine details. 
Extensive experiments show that \method demonstrates state-of-the-art performance on click-based interactive segmentation benchmarks, and shows competitive performances for various interaction formats.
In general, the contribution of this work could be summarized in three folds:
First, our work makes FocalClick~\cite{focalclick} benefit from large-scale training, thus reaching state-of-the-art performance for click-based segmentation benchmarks. 
Second, we design a decomposed pipeline that extends FocalClick for novel interaction forms like scribbles, boxes, and coarse masks.
Third, we carry out a thorough study of various interactions and formulate well-defined train/val protocols and benchmarks, which would benefit the community in exploring more types of interaction. 

\textbf{Difference from conference version.} This manuscript improves the conference version \cite{focalclick} significantly with wider extensions for up-scaled training and support for more interactions.  1)~In \cref{sec:scale_up}, we up-scaled the model design and training data for each stage of FocalClick and apply different training strategies in each stage.   
2)~We extend the user-interaction support for FocalClick from click to wider types, including scribble, box and coarse masks in \cref{sec:extension}. At the same time, we thoroughly investigate the evaluation protocol for these new interactions.
3)~We give a more detailed experimental analysis for the newly added part and scaling strategies in  \cref{sec:expxl}, and add more discussions and comparisons with more recent works.

\section{Related Work}
\label{sec:related_work}
\noindent \textbf{Interactive segmentation.~} Before the era of deep learning, researchers~\cite{rother2004grabcut,gulshan2010geodesic,grady2006random,kim2010nonparametric} take interactive segmentation as an optimization procedure.
DIOS~\cite{xu2016deep} first introduces deep learning into interactive segmentation by embedding positive and negative clicks into distance maps, and concatenating them with the original image as input. It formulates the primary pipeline and train/val protocol for click-based interactive segmentation. After this, \cite{li2018latentdiversity, liew2019multiseg} focus on the issue of ambiguity and predict multiple potential results and let a selection network or the user choose from them. FCANet~\cite{firstclick} emphasizes the particularity of the first click and uses it to construct visual attention. BRS~\cite{jang2019brs} first introduces online optimization, which enables the model to update during annotation. f-BRS~\cite{fbrs} speeds up the BRS~\cite{jang2019brs} by executing online optimization in specific layers.
CDNet~\cite{chen2021cdnet} introduces self-attention into interactive segmentation to predict more consistent results. RITM~\cite{sofiiuk2021ritm} add the previous mask as network input to make the prediction more robust and accurate. 
SimpleClick~\cite{liu2023simpleclick} leverage ViT encoders to extract more power features, achieving the state-of-the-art performance.
Besides click-based segmentation, other formats of interactions have also been explored. IOG~\cite{iog} proposes to combine the boxes and clicks together.  Other works~\cite{ScribbleSup, GRanking,bai2014error,GFilter,DeepIGeoS,IFIS, appearancesimilarity, bai2014error} investigate scribbles as scribbles are more flexible than clicks.

\noindent\textbf{Segment anything series.} SAM~\cite{sam} is a powerful model for segmentation. It is essentially designed for click-based interactive segmentation, and can further be extended to support automatic segmentation or using the box as a prompt. SAM serves as the foundation model for various downstream tasks across different visual tasks~\cite{li2023semanticsam, lin2023samus,cheng2023segmenttrack,yang2023trackanything,yang2023sam3d, zhang2023sam3d, wang2023captionsam, yu2023inpaintanything}. Some previous works make improvements to SAM. For example, SAM-HQ~\cite{hqsam} explores to refine the masks produced by SAM. SEEM~\cite{seem} unifies more modalities of input as tokens. Large numbers of works~\cite{SAM-Adapter,sam-Camouflage,sam-medical,zhang2023samadapter,chen2023masamadapter} design adapter layers or LoRAs to adjust SAM to specific tasks. 
However, the SAM series makes some compromises for automatic segmentation thus sacrificing the performance for interactive segmentation. 
In this work, we get inspiration from them to unleash the power of large-scale training and take a further stage for the specific task of interactive segmentation.

\noindent \textbf{Local inference in interactive segmentation.~}Some previous methods~\cite{fbrs,chen2021cdnet,sofiiuk2021ritm} \cite{fbrs,chen2021cdnet,sofiiuk2021ritm} crop the region around the predicted target for subsequent inference, similar to our Target Crop. However, since they generate final masks on these crops, they must maintain high resolution. In contrast, FocalClick utilizes the Target Crop to locate the Focus Crop without relying on the Segmentor for fine details, allowing us to resize it for higher speed. Other works~\cite{liew2017regional,hao2021edgeflow,forte2020getting99} refine predictions in a coarse-to-fine manner but incur high computational costs. RIS-Net~\cite{liew2017regional} refines using multiple ROIs based on click positions, while EdgeFlow~\cite{hao2021edgeflow} and 99\%AccuracyNet~\cite{forte2020getting99} focus on boundary refinements. Unlike these, FocalClick strategically selects local patches, significantly reducing FLOPs by decomposing the process into coarse segmentation and refinement, making interactive segmentation more efficient.

\begin{figure*}[t]
\newcommand{\image}{\includegraphics[width=2\columnwidth]}
\centering 
\image{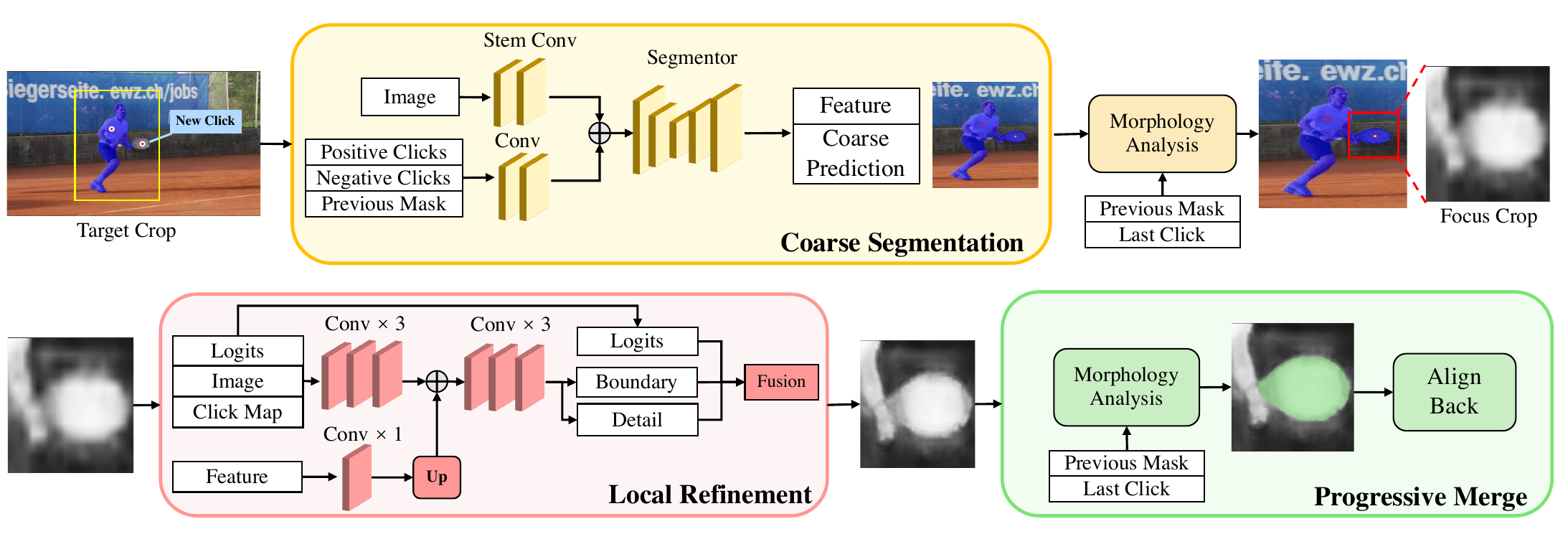} 
\vspace{-3mm}
\caption{ \textbf{Overall framework of FocalClick.} We take the image, two click maps, and the previous mask as input. We use binary disks with radius 2 to represent the click. First, we select the \textbf{Target Crop} around the target object and resize it to a small size. It is then sent into \textbf{Segmentor} to predict a coarse mask. Next, we chose a \textbf{Focus Crop} by calculating the different regions between the previous masks and the coarse prediction to refine the details. At last,  \textbf{Progressive Merge} updates the local part that the user intends to modify and preserves the details in other regions.
}
\label{fig:pipeline_original}
\vspace{-3mm}
\end{figure*}
\section{Method}
We begin by introducing our efficient FocalClick pipeline and elaborating on a newly proposed task, Interactive Mask Correction, along with our benchmark. Following this, we present the extensions and enhancements that define \method in the subsequent sections.

\subsection{FocalClick: Multi-stage Decoupled Pipeline}
The core idea of the FocalClick pipeline is to break down a single, heavy inference over the entire image into two lightweight predictions on smaller patches. As illustrated in Fig.~\ref{fig:pipeline_original}, the process begins with Target Crop, which selects a patch centered around the target object, resizes it to a smaller scale, and feeds it into the Segmentor to generate a coarse mask. Then, Focus Crop identifies a local region that requires refinement and sends the zoomed-in patch to the Refiner for further enhancement. Finally, Progressive Merge integrates these localized predictions back into the full-resolution mask. This iterative refinement process ensures that only a small local region is updated after each user interaction, while all pixels in the final prediction benefit from repeated refinements distributed across multiple rounds.

\noindent \textbf{Target crop.} The objective is to eliminate background information that is irrelevant to the target object. 
To achieve this, we first determine the minimum bounding box that encloses both the previous mask and the newly added click. This bounding box is then expanded by a factor of $r_{TC}=1.4$ following \cite{fbrs,sofiiuk2021ritm}. Afterward, we crop the relevant input tensors, including the image, previous mask, and click maps, and resize them to a smaller scale for efficient processing.

\noindent \textbf{Coarse segmentation.~} This step aims to generate an initial rough mask for the target object, which serves as a foundation for locating the Focus Crop and enabling further refinement. The Segmentor can be any segmentation network \cite{pspnet,chen2017deeplab,long2015fcn,xiao2018upernet,peng2017largekernel}, allowing customization for different scenarios. In our implementation, we adopt state-of-the-art methods such as HRNet+OCR \cite{wang2020hrnet,yuan2020ocr} and SegFormer \cite{xie2021segformer} as representative architectures. As illustrated in Fig.~\ref{fig:pipeline_original}, we follow the RITM framework \cite{sofiiuk2021ritm}, incorporating two convolutional layers to adjust the channel and scale of the click maps, followed by feature fusion after the stem layers.

\noindent \textbf{Focus crop.~}It aims to locate the area that the user intends to modify. We first compare the differences between the primitive segmentation results and the previous mask to get a Difference Mask $ {M_{xor}}$. We then calculate the max connected region of ${M_{xor}}$ that contains the new click, and we generate the external box for this max connected region.
Similar to the Target Crop, we make expansion with ratio $r_{FC}=1.4$. We note this region Focus Crop. Accordingly, we crop local patches on the input image and click maps. Besides, we use RoiAlign~\cite{he2017mask} to crop the feature and the output logits predicted by the Segmentor.

\noindent \textbf{Local refinement.~}It recovers the details of the coarse prediction in Focus Crop. 
We first extract the low-level feature from the cropped tensor using Xception convs~\cite{chollet2017xception}. At the same time, we adjust the channel number of the RoiAligned feature and fuse it with the extracted low-level feature. To get refined predictions, we utilize two heads to predict a Detail Map $M_d$ and a Boundary Map $M_b$, and calculate the refined prediction $M_r$ by updating the boundary region of the coarse predicted logits $M_l$, as in Eq.~\ref{eq:refine}.
\begin{equation}
\mathit{M_r  =  Sigmoid(M_b) * M_d + (1 - Sigmoid(M_b)) * M_l}
\label{eq:refine}
\end{equation}

\begin{figure*}[t]
\newcommand{\image}{\includegraphics[width=1.9\columnwidth]}
\centering 
\image{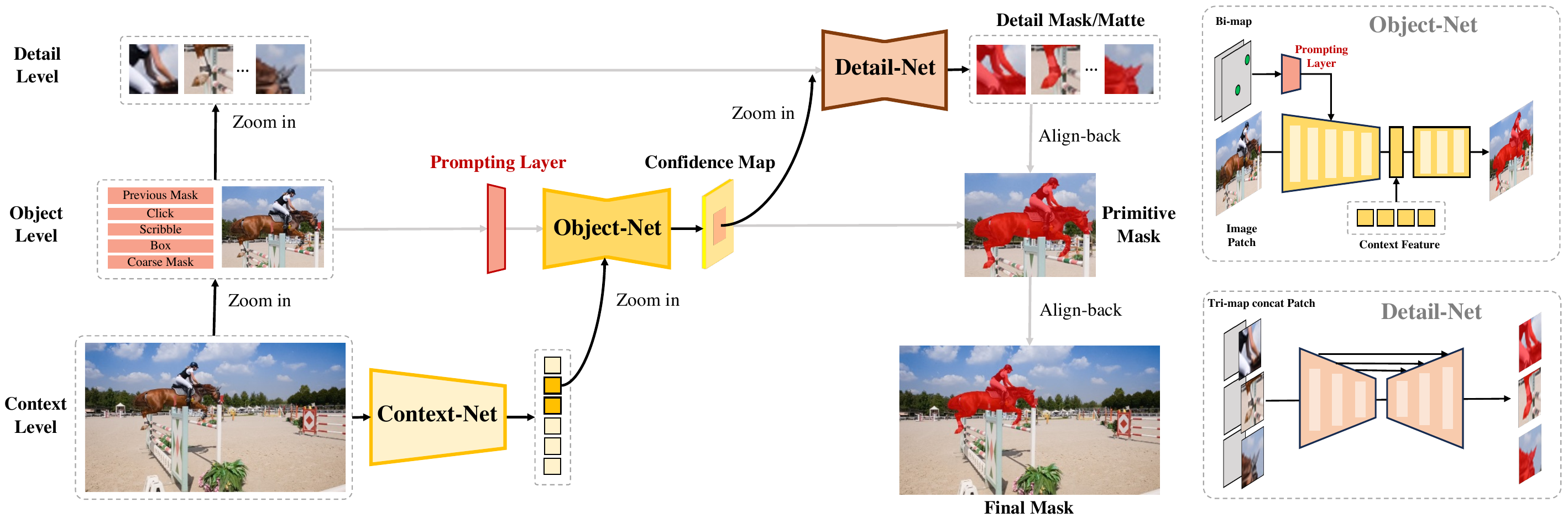} 
\vspace{-5pt}
\caption{\textbf{The extended framework of \method .}  The overall pipeline is decomposed into three stages that individually focus on context-, object-, and detail-level information. First, the full image is fed into the Context-Net to extract the global feature. Afterward, the global feature is zoomed in around the target object~(based on the previous segmentation results). The image patches around the object and the user-interaction embeddings are sent into the Object-Net to get a primitive mask. Next, we zoom in on the low-confidence local regions of the primitive masks into the Detail-Net to refine the prediction. At last, we align back the detail mask and the primitive mask to the full image.
}
\label{fig:pipeline}
\vspace{-5pt}
\end{figure*}

\noindent \textbf{Progressive merge.~}When annotating or editing masks, we do not expect the model to update the mask for all pixels after each click. Otherwise, the well-annotated details would be completely over-written. Instead, we only want to update in limited areas that we intend to modify.  
Similar to the method of calculating Focus Crop, Progressive Merge distinguishes the user intention using morphology analysis. After adding a user click, we simply binarize the newly predicted mask with a threshold of 0.5 and calculate a different region between the new prediction and the preexisting mask, then pick the max connected region that contains the new click as the update region~(the part the green in Fig.~\ref{fig:pipeline}).  In this region, we update the newly predicted mask onto the previous mask, and we keep the previous mask untouched in other regions.  

When starting with a preexisting mask or switching back from other segmentation tools, we apply Progressive Merge to preserve the correct details. While annotating from scratch, we active the progressive mode after 10 clicks.

\noindent \textbf{Training supervision.~}The supervision of the boundary map $M_b$ is computed by down-sampling the segmentation ground truth 8 times and resizing it back. The changed pixels could represent the region that needs more details. We supervise the boundary head with Binary Cross Entropy Loss~${L_{bce}}$. The coarse segmentation is supervised by Normalized Focal Loss~$L_{nfl}$ proposed in RITM~\cite{sofiiuk2021ritm}. For the refined prediction, we add boundary weight~(1.5) on the NFL loss, and we note it as $L_{bnfl}$, the total loss could be calculated as Eq.~\ref{eq:loss}.
\begin{equation}
\mathit{L = L_{bce} + L_{nfl} + L_{bnfl} }
\label{eq:loss}
\end{equation}

\subsection{ Interactive Mask Correction }
\label{sec:mask}
In practical applications, large portions of annotation tasks provide pre-inferred masks. In this case, annotators only need to make corrections on them instead of starting from zero. Besides, during annotation, when annotators switch to matting, lasso, or polygon tools and switchback, we also expect to preserve the pixels annotated by other tools. However, existing methods predict new values for all pixels after each click. Thus, they are not compatible with modifying preexisting masks or incorporating other tools.

To solve this problem, we make the following attempts:
1)~We construct a new benchmark, DAVIS-585, which provides initial masks to measure the ability of mask correction.
2)~We prove that our FocalClick shows significant superiority over other works in this new task.

\noindent \textbf{New benchmark:~DAVIS-585. }Existing works uses GrabCut~\cite{rother2004grabcut}, Berkeley~\cite{berkeley}, DAVIS~\cite{davis}, SBD~\cite{SBD} to evaluate the performance of click-based interactive segmentation. However, none of them provides initial masks to measure the ability of Interactive Mask Correction. 
Besides, GrabCut and Berkeley only contain 50 and 100 easy examples, making the results not convincing. SBD provides 2802 test images, but they are annotated by polygons with low quality. DAVIS is firstly introduced into interactive segmentation in \cite{li2018latentdiversity}. It contains 345 high-quality masks for diverse scenarios. However, as \cite{li2018latentdiversity} follows the setting of DAVIS2016, it merges all objects into one mask. Hence, it does not contain small objects, occluded objects, and no-salient objects. In this paper, we chose to build a new test set based on DAVIS~\cite{davis} for its high annotation quality and diversity, and we made two modifications:

First, we follow DAVIS2017 which annotates each object or accessory separately, making this dataset more challenging.  We uniformly sample 10 images per video for 30 validation videos, and we take different object annotations as independent samples. Then we filter out the masks under 300 pixels and finally get 585 test samples, so we call our new benchmark \textbf{DAVIS-585}.

Second, to generate the flawed initial masks, we compare two strategies:~1)~Simulating the defects on ground truth masks using super-pixels. 2)~Generating defective masks using offline models. We find the first strategy has two advantages: 1)~It could control the distribution of error type and the initial IOUs. 2)~The simulated masks could be used to measure the ability to preserve the correct part of preexisting masks. 
Therefore, we use super-pixels algorithm\footnote{https://github.com/Algy/fast-slic} to simulate the defect.  We first use mask erosion and dilation to extract the boundary region of the ground truth mask. We then define three types of defects: boundary error, external FP~(False positive), and internal TN~(True Negative).
After observing the error distribution in real tasks,  we set the probability of these three error types to be [0.65, 0.25, 0.1] and follow Alg.~\ref{alg1} to control the quality of each defective mask. To decide the quality range, we carry out a user study and find that users intend to discard the given masks when they have IOU lower than 75\%. Considering that current benchmarks use NoC85~(Number of Clicks required on average to reach IOU 85\%) as the metric, we control our simulated masks to have IOUs between 75\% and 85\%. \\

\begin{algorithm}[t]
    \small
	\caption{Simulate Defective Mask using Super-Pixels} 
	\label{alg1} 
	\small
	\begin{algorithmic}
		\REQUIRE Image, GTMask, maxIOU=0.85, minIOU=0.75
		\STATE SimMask $\gets$ GT
		\WHILE{ $\mathrm{True} $ } 
		\STATE ErrorType $\gets$ Rand([Boundary, External, Internal])
		\STATE PixelNumber $\gets$ Rand([50, 100, 200, 300, 500, 700])
		\STATE SuperPixels $\gets$ Slic(Image, PixelNumber)
		\STATE SimMask $\gets$ Merge(~ErrorType, SuperPixels, SimMask)
		\STATE MaskIOU $\gets$ IOU(~SimMask, GTMask)
		
		\IF{MaskIOU $<$ minIOU} 
		\STATE Break
		\ELSIF{MaskIOU $>$ maxIOU}
		\STATE Continue 
		\ELSE
		\STATE Return SimMask
		\ENDIF 
		\ENDWHILE 
	\end{algorithmic}
\end{algorithm}
\vspace{-3mm}

\section{Scaling up the multi-stage pipeline}
\label{sec:scale_up}
Following the multi-stage design of FocalClick, we scale up the model structure and training data of each stage and further propose a stronger version, termed \method.

The framework of \method is depicted in \cref{fig:pipeline}. \method decomposes the pipeline for interactive segmentation into three stages to model the features in context-, object-, and detail-level individually. The full image is sent into the Context-Net to extract global features. We crop the target region~(based on the previous round segmentation result) and feed the target-object patch with interaction embeddings into the Object-Net, this part takes charge of locating the target region indicated by the user interactions and produces a primitive mask. Based on the primitive mask, we further conduct another zoom-in around the low-confidence region and pass the small patches into the Detail-Net to produce high-quality results. At last, the refined predictions and the primitive masks are aligned back to the full image to get the final prediction.

\subsection{Structure Decomposition}
\label{sec:decomposed}
We introduce the detailed model structures of the FocalClick-XL. The pipeline is split into three meta-steps to focus on the information from context-, object-, and detail-level.

\noindent \textbf{Context-Net. } The main objective of the Context-Net is to understand complicated scenes and distinguish the potential objects from the background.
In the design, the Context-Net harnesses the weights from the SAM Encoder to capitalize on the robust ability of scene understanding and object priors.
To adapt the SAM encoder, we integrate a learnable adapter after each transformer layer. Our adapter layers are designed as bottlenecks consisting of ``Conv-GeLU-Conv''.  The convolutional layers reduce the channels of the feature map and the GeLU activation introduces the non-linearities. 
Commencing from the second round of interactions, we compute the Region of Interest (RoI) for the target object using predicted masks from the previous round. Subsequently, we employ RoI-Align~\cite{he2017maskrcnn} to crop the RoI from the SAM features, preparing for feature fusion for the Object-Net.

\noindent \textbf{Object-Net.} It encodes the provided interactions to segment the user-intended parts. The structure follows SegFormer~\cite{xie2021segformer}. The Object-Net receives the zoomed-in region of the full image and interaction maps as inputs. It extracts target-centric features endowed with intricate details and facilitates early fusion between interaction maps and the image. This early fusion significantly amplifies the model's controllability in modifying the fined details with excessive user interactions. 
The Object-Net also receives the RoI-Aligned features from the Context-Net, we perform an elementwise addition to fuse the target-centric feature extracted by the Object-Net with the RoI-aligned feature from Context-Net. This fused feature aims to encapsulate both global and local perspectives. 
Subsequently, we decode the fused feature for the primitive mask prediction. Specifically, a multi-layer perceptron (MLP) is employed to generate a one-channel binary mask. Notably, this primitive mask has already exhibited competitive performance across various datasets.
Throughout the training phase, we utilize normalized focal loss to supervise the primitive mask for click-based interactive segmentation. The Object-Net and the adapter layers of the Context-Net are optimized together on object-centric samples.

\noindent \textbf{Detail-Net.} This module further refines the low confidence of the primitive mask of the Object-Net to produce high-quality masks.
The Detail-Net is designed as a universal module that could act as a plug-in to support different user interactions. Inspired by image matting methods~\cite{hou2019contextmatting,xu2017deepimagematting}, Detail-Net takes a ``tri-map'' along with the image as input. The tri-map consists of a high-confidence foreground mask,  a high-confidence background mask, and a mask for the low-confidence regions. 
The positive/negative user interactions and the high-confidence part of the primitive mask predicted by the Object-Net are merged to act as the tri-maps. After zooming in on the target object for Object-Net, the input of the Detail-Net is magnified again around the uncertain region to capture more details. 
The structure of Detail-Net is a U-shape network with a MobileNetV2~\cite{sandler2018mobilenetv2} as the encoder. 
We collect large numbers of data from both the segmentation dataset~\cite{lin2014coco,gupta2019lvis,borji2015msra10k,ade20k,wang2017learningduts,xu2018youtubevos,thinobject,cong2020dovenet} and the image matting dataset~\cite{li2022bridgingam2k,xu2017deepimagematting,qiao2020distinct646} to train the Detail-Net. Use leverage MSE Loss to force Detail-Net to predict the transparency value of the object. Thus, by forcing Detail-Net to learn the alpha value for the transparent parts like the hairs and fur, it is equipped with the ability to capture finer details by producing the alpha-matte. We could simply use a threshold of 0.5 to binarize the prediction to give segmentation masks.

\subsection{Progressive Magnification}
\label{sec:magnifier} 
The cascaded subnets bring extra burden for the computation and inference speed. To mitigate this problem, we propose a progressive magnification strategy to increase the efficiency of our pipeline.

First, we make the heaviest Context-Net as an ``offline'' extractor. As the Context-Net receives the full image as input, it models each pixel equally thus irrelevant to the user interaction. In our pipeline, the  Context-Net is executed only once per image and could be pre-executed before receiving user interactions. Adhering to the original settings, the full image is resized to $1024\times1024$ as input.

Second, we progressively zoom in on the local regions to support small-size input for the Object-Net and Detail-Net. Concretely, based on the segmentation results of the previous round, we calculate and expand the box around the segmentation target to get the input of the Object-Net. If the previous around the mask is void, we calculate the box covering the full image.
Considering box region is zoomed in, we do not need the high input resolution to capture the details. Thus, the input size of Object-Net is set as $384\times384$ for better efficiency. 
A similar zoom strategy is also applied to the Detail-Net, we crop the small local regions around the low-confidence pixels and formulate multiple patches as a batch. Thus, although the input size of Detail-Net is set as  $256\times256$, we could still get highly refined details.
In this way, we make each subnet focus on a different part of the images, effectively reducing the computation burden and inference time of our FocalClick-XL pipeline.

\section{Extension to more interactions}
\label{sec:extension}
We further present the extension for more forms of user interactions with a unified structure. At the same time, we investigate the evaluation protocol for different interactions.

\subsection{Prompting Unified Interaction}
\label{sec: interaction} 
\method supports various user interactions like click, box, scribble, coarse mask, \textit{etc.} In this section, we introduce the representation of different interaction forms and the transferable training scheme. In addition, we also elaborate on the user-interaction simulation strategies for constructing training samples.

\noindent \textbf{Interaction representation. } 
Different from SAM~\cite{sam} series that use coordinates to represent interactions, we represent different interactions uniformly with a 2-channel ``bi-map'' for the positive and negative signals. For clicks, we follow previous works~\cite{sofiiuk2021ritm,focalclick,liu2023simpleclick} to encode clicks as round disks. Scribbles are directly divided into a positive and negative scribble map. Boxes are represented with a binary rectangle mask. As for coarse masks, the mask would be regarded as the positive interaction channel, and an all-zero negative map is represented.  In the following part, we first present the training strategies for supporting various of interactions, afterward, we present how to simulate diversified user interactions during training. 

\noindent \textbf{Promptable transferation.} As we formulate different user interactions as a unified 2-channel map, \method can deal with various tasks using the same architecture.  We concatenate the 2-channel map with the input image and use the Prompting layer~(projection conv) to encode interaction signals.
Considering clicks could be regarded as the basic elements for different interactions, we first train \method on clicks as the pertaining.
Specifically, we pre-train the Object-Net and the adapter layers of the Context-Net together on simulated clicks and object-centric data~(Context-Net takes the full image, and we sample boxes around the GT masks for the input of Object-Net).  After pretraining, 
we find that tuning only the Promtping Layer is sufficient for transferring \method to different tasks. The Prompting Layer of the Object-Net projects the interaction maps into the control signals. After task-specific tuning, the control signals of different forms of interactions could be unified into the same feature space. Given a new interaction definition, \method could be easily transferred to new tasks with low training and storage requirements.

\begin{figure}[t]
\centering
\includegraphics[width=1.0\linewidth]{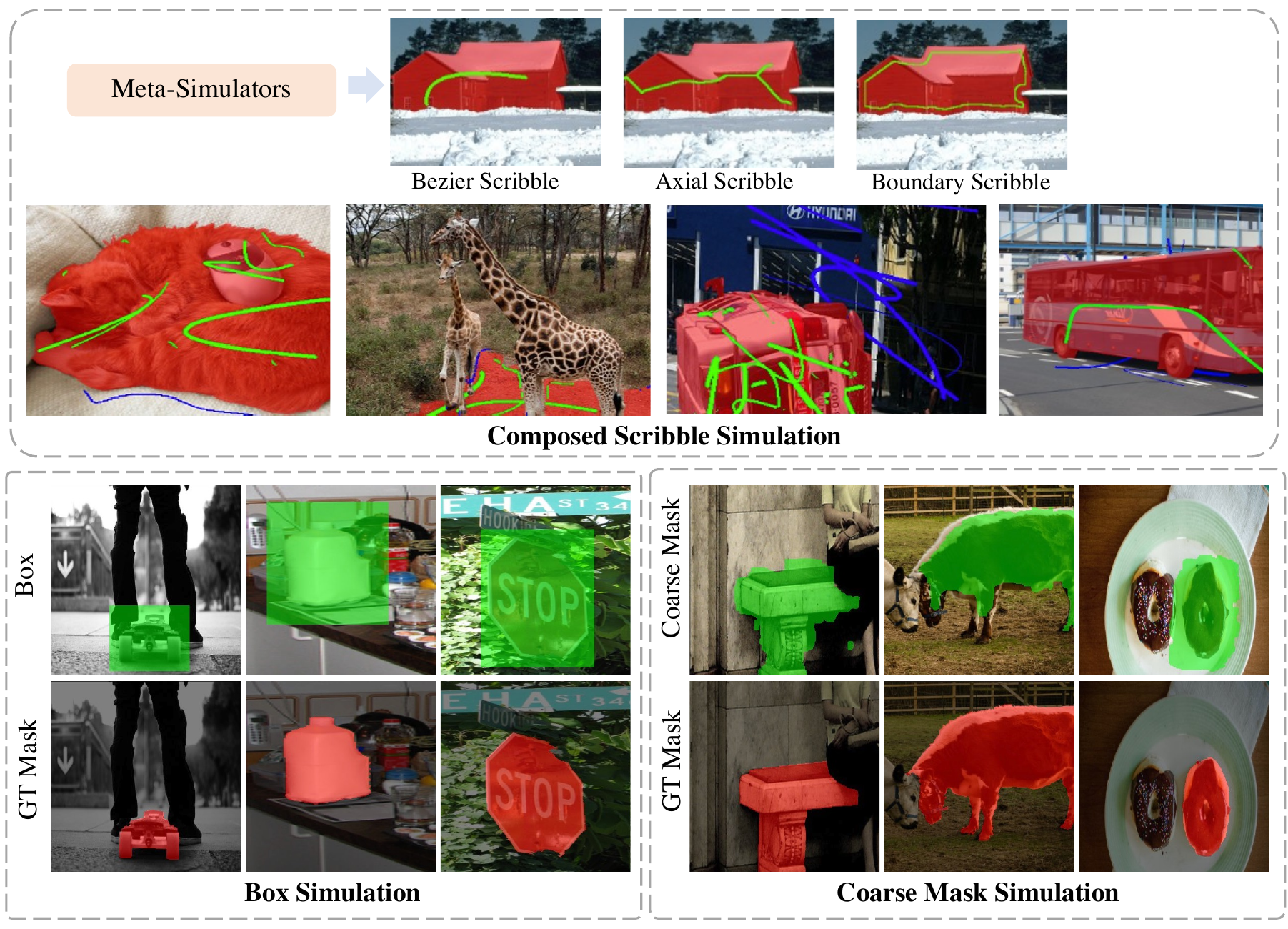} 
\vspace{-20pt}
\caption{%
    \textbf{Simulated user interactions}, such as scribbles, boxes, and coarse masks. For the scribbles, we first develop several meta-simulators and compose them for more versatile results.
}
\label{fig:interactionmap}
\end{figure}

\noindent \textbf{Interaction simulation. }  The strategies used to simulate user interactions play a pivotal role in model training. To simulate clicks, we adopt methodologies from previous works~\cite{sofiiuk2021ritm, focalclick, liu2023simpleclick, mahadevan2018iteratively}, placing clicks iteratively within the maximin-error regions identified in current predictions. Additionally, we assign higher probabilities to sample clicks around boundary regions. For boxes, we derive the minimum bounding box from ground truth masks and introduce variations in box sizes and locations. Coarse masks undergo random erosion, dilation, and downsampling processes to augment ground truth masks. Moreover, we incorporate current model predictions into this augmentation process. Visual examples of these strategies are illustrated in \cref{fig:interactionmap}.

Simulating human scribbles poses the greatest challenge due to their flexible and arbitrary nature in shape. Past approaches~\cite{appearancesimilarity, IFIS} have utilized simplistic strategies such as linking points or filling basic geometries. This study introduces multiple meta-simulators designed to generate diverse scribbles. As demonstrated in \cref{fig:interactionmap}, the bezier scribble uses the bezier function to draw curves within the mask regions; the axial scribble calculates the media axis of the given mask; the boundary scribble draws lines along with the mask boundary. For the stroke thickness, we randomly choose values from 3 to 7. We combine these four strategies to generate diversified scribbles to simulate the diversified user input in real practices.

\subsection{Evaluation Protocols}
\label{sec:evaluation} 
Most of the previous works investigate the click-based setting. There are no existing standard evaluation protocols for other interactions like scribbles and coarse masks. Thus, it is not trivial to build a benchmark for evaluation.
We follow the evaluation protocol of click-based methods and extend it to scribbles. Besides, we formulate the protocols for boxes and coarse masks. 
 
\begin{figure}[t]
\centering 
\includegraphics[width=0.99\linewidth]{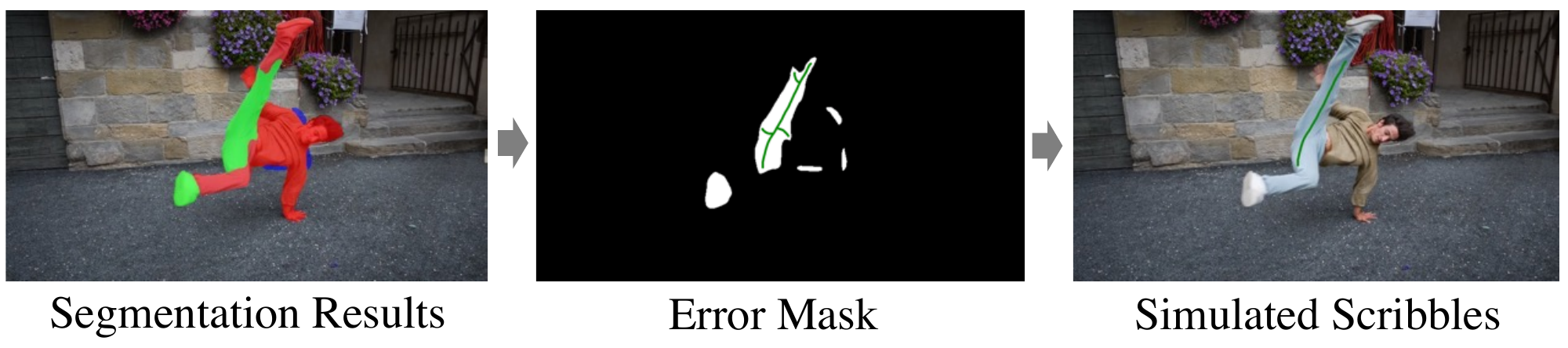} 
\vspace{-5pt}
\caption{ \textbf{The procedure of deterministic scribble generation.} The true positives, false negatives, and false positives of the segmentation result are marked in red, green, and blue respectively. We first compute the largest connected region of the error mask~(middle image) and generate deterministic scribbles as the following interaction. }
\label{fig:scr_eval}
\end{figure}

\begin{algorithm}[t]
    \small
    \caption{Deterministic Scribble Simulator}
    \label{code}
    \begin{algorithmic}[1]
        \STATE $max\_mask \gets \max(error\_mask)$ 
        \STATE $skel\_mask \gets$ MedialAxis$(max\_mask)$
        \STATE $Graph \gets$ RadiusNeighbourGraph$(skel\_mask)$
        \FOR{$subgraph \in$ Connected$(Graph)$}
            \WHILE{True}
                \STATE $cycle \gets$ FindCycle$(subgraph)$
                \IF{$cycle == \text{None}$} 
                    \STATE \textbf{break}
                \ELSE
                    \STATE RemoveCycle$(subgraph, cycle)$
                \ENDIF
            \ENDWHILE
        \ENDFOR
        \STATE $distance \gets [~]$
        \FOR{$v \in$ Graph.nodes()}
            \STATE $max\_path \gets$ ShortestPath$(Graph, v)$
            \STATE distance.append$(max\_path)$
        \ENDFOR
        \STATE $longest\_path \gets \max(distance)$
        \STATE $scribble \gets$ BezierCurve$(longest\_path)$
    \end{algorithmic}
\end{algorithm}

\noindent \textbf{Scribble-based evaluation.}  We extend the click-based protocol for scribbles. The challenge is that clicks could simply be added at the center of the error region, but not for scribbles, as they have various shapes, which introduces randomness. 
Accordingly, we utilize a deterministic scribble simulator that synthesizes scribbles according to the shape and size of the given mask. 
As demonstrated in \cref{fig:scr_eval}, similar to the click-based protocol, we first calculate the max error regions.  Then, the medial axis for the largest error mask is computed to obtain the skeleton of the objects. Afterward, we transform the skeleton mask into a radius neighbor graph, where the neighborhood of a vertex points at a distance less than the radius from it. Then we divide the graph into connected components sub-graphs and remove its cycles. Finally, we will create a Bezier curve with the points in the graph's longest path.  The pseudo-code is shown in Alg.~\ref{code}.
Therefore, we could iteratively add scribbles on the FP or FN regions. Thus, we generalize the NoC metric for clicks to NoS~(Number of Scribbles), and report NoS85/90 and NoF$^{20}$85/90.

\noindent \textbf{Box and coarse mask evaluation.} Box and coarse masks only provide single-round interactions. For the boxes, we measure the mIoU. As for the coarse masks, we exert different levels of perturbations on the ground truth masks to get a series of coarse mask datasets. We report the mIoU for the predicted masks with those coarse mask datasets to verify the robustness of our method.

Specifically, to evaluate \method's ability for coarse mask refinement. We design a mask perturbation strategy to generate flowed masks with different initial IoUs.  Specifically, we leverage interactive perturbation by applying different types of kernels to conduct erosion or dilation on the ground truth masks.  For each interaction, we add a small perturbation on the current mask, until it reaches the target IoU range.  Thus we generate different levels of coarse masks with ranges of [0.85, 0.90], [0.75, 0.80], [0.65, 0.70], [0.55, 0.60], [0.45, 0.50].

\section{Experiments for FocalClick} \label{sec:exp}

This section introduces the basic experiment results of  FocalClick. We start with the experiment configurations, then we introduce the ablation studies to verify the effectiveness of each component and report the performance of interactive mask correction.
Afterwards, in the next section, we further analyze the extended part of \method.

\subsection{Experimental Configuration}
\noindent \textbf{Model series.~}To satisfy the requirement for different scenarios, we design two versions of models with different input resolutions as demonstrated in \cref{tab:series}. The S1 version is adapted to edge devices and the plugins for web browsers.  The S2 version would be suitable for CPU laptops. In this paper, we conduct experiments for both SegFormer~\cite{xie2021segformer} and HRNet~\cite{wang2020hrnet} as our Segmentor to show the universality of our pipeline. In the rest of the paper, we use S1, S2 to denote different versions of our model.

\begin{table}[t]
\begin{center}
\caption{ \textbf{Model configurations} for the basic FocalClick. We present two versions with different resolutions.
}
\label{tab:series}
\vspace{-5pt}
\begin{tabular}{l|c|c  }
\toprule[1pt]
Model Series       & Segmentor Input &  Refiner Input \\
\hline
FocalClick-S1  & $128\times128$ & $256\times256$ \\
FocalClick-S2  & $256\times256$ & $256\times256$ \\
\bottomrule[1pt]
\end{tabular}
\end{center}
\vspace{-15pt}
\end{table}

\noindent \textbf{Training protocol.~}We simulate the Target Crop by randomly cropping a region with a size of $256\times256$ on the original images. Then, we simulate the Focus Crop by calculating the external box for the ground truth mask, or making random local crops centered on boundaries with the length of 0.2 to 0.5 of the object length. 
Then, we add augmentations of random expansion from 1.1 to 2.0 on the simulated Focus Crop. Thus, the whole pipeline of Segmentor and Refiner is trained together end-to-end.

For the strategy of click simulation, we exploit iterative training~\cite{mahadevan2018iteratively} following RITM~\cite{sofiiuk2021ritm}. Besides the iteratively added clicks, the initial clicks are samples inside/outside the ground truth masks randomly following \cite{xu2016deep}. The maximum number of positive~/negative clicks is set as 24 with a probability decay of 0.8. 
For the hyper-parameters, following RITM~\cite{sofiiuk2021ritm}, we train our models on a combination dataset of COCO~\cite{lin2014coco} and LVIS~\cite{gupta2019lvis}. We also report the performance of our model trained on SBD~\cite{SBD}, and a large combined dataset~\cite{lin2014coco,gupta2019lvis,borji2015msra10k,ade20k,wang2017learningduts,xu2018youtubevos,thinobject,cong2020dovenet}.
During training, we only use flip and random resize with the scale from 0.75 to 1.4 as data augmentation.  We apply Adam optimizer of $\beta_1 = 0.9, \beta_1 = 0.999$. We denote 30000 images as an epoch and train our models with 230 epochs. We use the initial learning rate as $5\times 10^{-4}$ and decrease it 10 times at the epoch of 190 and 220. We train each of our models on two V100 GPUs with batch size 32. The training takes around 24 hours. \\

\noindent \textbf{Evaluation protocol.~}We follow previous works~\cite{xu2016deep,chen2021cdnet,sofiiuk2021ritm,jang2019brs,fbrs,firstclick} to make fair comparisons. During evaluation, the clicks are automatically simulated with a fixed strategy:  Each click would be placed at the center of the largest error region between the previous prediction and the ground truth mask. For example, when starting from scratch, the first click would be placed at the center of the ground truth mask. Additional clicks would be added iteratively until the prediction reaches the target IOU~(Intersection over Union) or the click number reaches the upper bound.    

For the metrics, we report NoC~IOU~(Number of Clicks), which means the average click number required to reach the target IOU. Following previous works, the default upper bound for click number is 20. We note the sample as a failure if the model fails to reach the target IOU within 20 clicks. Hence, we also report NoF~IOU~(Numbers of Failures) to measure the average number of failures.

\subsection{ Ablation Study}
We conduct plenty of ablation studies for each module of FocalClick and report experimental results on both the original DAVIS  and our DAVIS-585 dataset. \\

\begin{table}[t]
\small
\caption{ \textbf{Ablation studies for FocalClick} on both interactive segmentation from scratch and interactive mask correction. `TC', `FC', `PM' denote Target Crop, Focus Crop, and Progressive Merge. `NoC', `NoF' stand for the Number of Clicks and the Number of Failures. }
\vspace{-10pt}
\begin{center}
\scalebox{0.75}{
\begin{tabular}{ll|c|c|c|c|c|c}
\toprule[1pt]
\multicolumn{2}{l|}{} & \multicolumn{3}{c|}{DAVIS} & \multicolumn{3}{c}{DAVIS-585}   \\
\cline{3-8}
\multicolumn{2}{l|}{Method } & NoC85 & NoC90 &  NoF90 & NoC85 & NoC90  & NoF90 \\
\hline
\multicolumn{2}{l|}{Naive-B0-S1} &  10.70 & 15.60 & 250 & 12.26 & 15.99 & 441  \\
\multicolumn{2}{l|}{+ TC} &  5.70 & 9.56 & 119 & 5.84 & 9.45 & 184  \\
\multicolumn{2}{l|}{+ TC+ FC} &  5.15 & 7.66 & 72 & 5.41 & 8.52 & 145  \\
\multicolumn{2}{l|}{+ TC+ FC+ PM} &  5.13 & 7.42 & 64 & 2.63 & 3.69 & 54  \\
\hline
\multicolumn{2}{l|}{Naive-B0-S2} &  5.24 & 9.74 & 129 & 7.00 & 10.81 & 251  \\
\multicolumn{2}{l|}{+ TC} &  4.52 & 5.86 & 58 & 4.02 & 6.53 & 99 \\
\multicolumn{2}{l|}{+ TC+ FC} &  4.15 & 5.55 & 56 & 3.94 & 6.23 & 93  \\
\multicolumn{2}{l|}{+ TC+ FC+ PM} &  4.04 & 5.49 & 55 & 2.21 & 3.08 & 41  \\
\bottomrule[1pt]
\end{tabular}}
\end{center}
\label{tab:ablation}
\end{table}

\noindent \textbf{Holistic analysis.~}We verify the effectiveness for each of our novel components in \cref{tab:ablation}.  We first construct a naive baseline model based on SegFormer-B0, noted as Naive-B0-S1/S2. It takes the full image as input and does not apply TC~(Target Crop), FC~(Focus Crop), and PM~(Progressive Merge), which is similar to early works like \cite{xu2016deep,jang2019brs,li2018latentdiversity}.  It is shown that this kind of pipeline performs poorly, especially for small input resolutions S1~($128\times128$). Most test samples fail to reach the target IOU within 20 clicks.  Then, after progressively adding the TC, FC, and PM, we observe that each component brings steady improvement for annotating from both initial masks and scratch. 

Making comparisons between S1 and S2, we find that the naive version heavily relies on the input scale. The performance drops tremendously from S2 to S1. However, with the assistance of TC, FC, and PM, the disadvantage of small input could be compensated. \\

\noindent \textbf{Analysis for cropping strategy.~}We first count the average area of the Focus Crop, Target Crop and calculate the ratio to full image in \cref{tab:area}. It shows that our cropping strategy is effective in selecting and zoom-in the local regions.

\begin{table}[t]
\small
\caption{ \textbf{Statistics for the area} of Focus Crop and Target Crop. We report the ratio relative to the full scale image.
}
\vspace{-10pt}
\begin{center}
\scalebox{0.75}{
\begin{threeparttable}
\begin{tabular}{l|c|c|c|c|c  }
\toprule[1pt]
    Ratio to Image    &  GrabCut &  Berkeley &  SBD  & DAVIS  & DAVIS-585 \\
\hline
Focus Crop  & 54.15\% & 31.17\% & 10.15\% & 11.6\% & 8.76\%\\
Target Crop  & 89.34\% & 68.81\% & 27.56\% & 40.50\% & 28.93\%\\
\bottomrule[1pt]
\end{tabular}
\end{threeparttable}
}
\end{center}
\vspace{-2mm} 
\label{tab:area}
\end{table}

In \cref{tab:refiner}, we verify the robustness of our cropping strategy. The results show that the fluctuation caused by the hyper-parameter is negligible compared with the improvement brought by the modules. Besides, for the evaluation result in \cref{tab:interDAVIS},  we simply set those ratios as 1.4 following previous works~\cite{fbrs,sofiiuk2021ritm,chen2021cdnet}. However, \cref{tab:refiner} shows that our work could reach even higher performance with a more delicate tuning.

\begin{figure}[t]
\newcommand{\image}
{\includegraphics[height=2.1cm]}
\centering 
\tabcolsep=0.07cm
\renewcommand{\arraystretch}{0.06}
\begin{tabular}{ccc}
\vspace{1mm}
\image{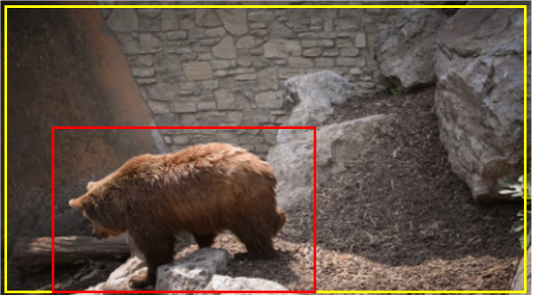} &
\image{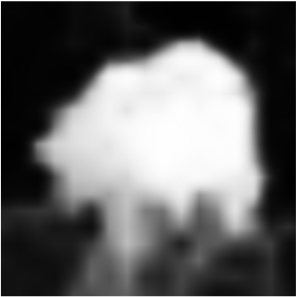} &
\image{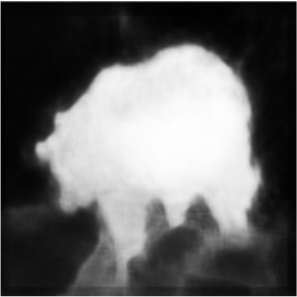} \\
\vspace{1mm}
\image{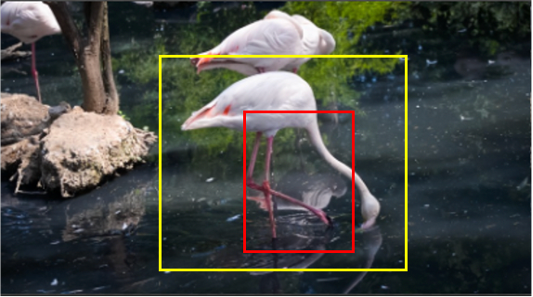} &
\image{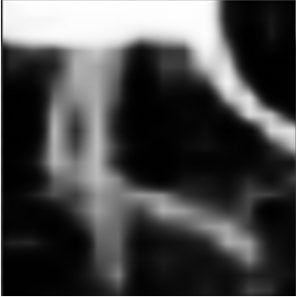} &
\image{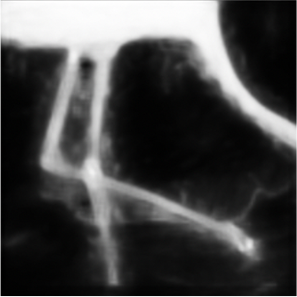} \\
\vspace{1mm}
\end{tabular}
\vspace{-10pt}
\caption{ \textbf{Qualitative analysis} for the effectiveness of Refiner. The first column denotes the Target Crop in yellow and the Focus Crop in red. The second and the third column demonstrate the mask in focus crop before and after refinement.
}
\label{fig:refiner}
\end{figure}

We also visualize the intermediate results of the Refiner to demonstrate its effectiveness. In \cref{fig:refiner}, the red boxes in the first column show the region selected by Focus Crop. The yellow box denotes the Target Crop~(The first row shows the case of the first click; hence the Target Crop corresponds to the entire image).
The second and the third column show the prediction results of the Segmentor and the Refiner. It demonstrates that Refiner is crucial for recovering the fine details.

\definecolor{darkseagreen}{rgb}{0.56, 0.74, 0.56}
\definecolor{lightpink}{rgb}{1.0, 0.71, 0.76}
\begin{table}[t]
\caption{ \textbf{The expand ratio} for TC~(Target Crop) and FC~(Focus Crop). The values show the NoC80/90 on DAVIS. The last row/column shows the performance without FC/TC.
}
\vspace{-10pt}
\begin{center}
\scalebox{0.75}{
\begin{threeparttable}
\begin{tabular}{l|c|c|c|c|c  }
\toprule[1pt]
\diagbox{ratio\_{FC}}{ratio\_{TC}}        &  1.2 &  1.4 &  1.6  & 1.8  & w/o TC \\
\hline
1.2  & \cellcolor{lightpink}5.15/7.21 & \cellcolor{lightpink}5.16/7.50 & \cellcolor{lightpink}5.23/7.90 & \cellcolor{lightpink}5.37/8.21 & 6.37/10.49\\
1.4  & \cellcolor{lightpink}5.07/7.10  & \cellcolor{lightpink}5.13/7.42 & \cellcolor{lightpink}5.15/7.77 & \cellcolor{lightpink}5.30/8.23 & 6.19/10.32\\
1.6  & \cellcolor{lightpink}4.99/7.07 & \cellcolor{lightpink}5.10/7.31 & \cellcolor{lightpink}5.11/7.80 & \cellcolor{lightpink}5.20/8.17 & 6.26/10.25\\
1.8  & \cellcolor{lightpink}4.99/6.95 & \cellcolor{lightpink}5.07/7.33 & \cellcolor{lightpink}5.08/7.64 & \cellcolor{lightpink}5.28/7.96 & 6.26/10.19\\
w/o FC  & 5.56/9.03 & 5.70/9.56 & 6.01/10.47 &6.56/11.31 &  \cellcolor{darkseagreen} 10.70/15.60\\
\bottomrule[1pt]
\end{tabular}
\end{threeparttable}
}
\end{center}
\vspace{-3mm}
\label{tab:refiner}
\end{table}

\begin{table*}[t]
\small
\caption{ \textbf{Quantitative results on DAVIS-585 benchmark}. The metrics `NoC' and `NoF' mean the average Number of Clicks required and the Number of Failure examples for the target IOU. All models are trained on COCO~\cite{lin2014coco}+LVIS~\cite{gupta2019lvis}.}
\vspace{-10pt}
\begin{center}
\scalebox{0.85}{
\begin{tabular}{ll|c|c|c|c|c|c|c|c|c|c|c|c}
\toprule[1pt]
\multicolumn{2}{l|}{} & \multicolumn{6}{c|}{From Initial Mask} & \multicolumn{6}{c}{From Scratch}   \\
\cline{3-14}
\multicolumn{2}{l|}{Method } & NoC85 & NoC90 & NoC95 & NoF85 & NoF90 & NoF95 & NoC85 & NoC90 & NoC95 & NoF85 & NoF90 & NoF95 \\

\hline
\multicolumn{2}{l|}{RITM-hrnet18s~\cite{sofiiuk2021ritm}} &  3.71 & 5.96 & 11.83 & 49 & 80 & 235 &        5.34 & 7.57 & 12.94 & 52 & 91 & 257  \\
\multicolumn{2}{l|}{RITM-hrnet32~\cite{sofiiuk2021ritm}} &  3.68 & 5.57 & 11.35 & 46 & 75 & 214 &         4.74 & 6.74 & 12.09 & 45 & 80 & 230  \\
\hline
\multicolumn{2}{l|}{Ours-hrnet18s-S1} &  2.72 & 3.82 & 5.86 & 37 & 57 & 97 & 5.62 & 8.08 & 13.73 & 53 & 98 & 274 \\
\multicolumn{2}{l|}{Ours-hrnet18s-S2} & 2.48 & 3.34 & 5.18 & 31 & 43 & 79 & 4.93 & 6.87 & 11.97 & 49 & 77 & 239  \\
\multicolumn{2}{l|}{Ours-hrnet32-S2} & 2.32 & 3.09 & 4.94 & 28 & 41 & 74 & 4.77 & 6.84 & 11.90 & 48 & 76 & 241  \\
\multicolumn{2}{l|}{Ours-segformerB0-S1} &  2.63 & 3.69 & 6.08 & 38 & 54 & 104 &  6.21 & 9.06 & 14.81 & 64 & 127 & 315  \\
\multicolumn{2}{l|}{Ours-segformerB0-S2} &  2.20 & 3.08 & 4.82 & 27 & 39 & 68 & 4.99 & 7.13 & 12.65 & 50 & 86 & 260   \\
\multicolumn{2}{l|}{Ours-segformerB3-S2} &  2.00 & 2.76 & 4.30 & 22 & 35 & 53 & 4.06 & 5.89 & 11.12 & 43 & 74 & 218   \\
\bottomrule[1pt]
\end{tabular}
}
\end{center}
\label{tab:interDAVIS}
\vspace{-10pt}
\end{table*}

\noindent \textbf{Computation analysis.~}The objective of FocalClick is to propose a practical method for mask annotation; efficiency is a significant factor. In \cref{tab:computation}, we make a detailed analysis and comparison for the number of parameters, FLOPs, and the inference speed on CPUs. We summarize the methods of the predecessors into five prototypes according to their backbone and input size.

\begin{table}[t]
\small
\caption{\textbf{Computation analysis} for FocalClick and classical pipelines. `B0/3' is short for SegFormer-B0/3. `Seg' denotes Segmentor, `Ref' denotes Refiner. `400', `600', `512' denote the default input size required by different models. The speed is measured on a CPU laptop with 2.4 GHz, 4$\times$Intel Core i5.  }
\vspace{-10pt}
\begin{center}
\scalebox{0.8}{
\begin{tabular}{ll|c|c|c|c|c|c}
\toprule[1pt]
\multicolumn{2}{l|}{} & \multicolumn{2}{c|}{Params/MB} & \multicolumn{2}{c|}{FLOPs/G} & \multicolumn{2}{c}{Speed/ms}  \\
\cline{3-8}
\multicolumn{2}{l|}{Base Model } & Seg & Ref & Seg & Ref   & Seg & Ref \\
\hline
\multicolumn{2}{l|}{hrnet18s-400~\cite{sofiiuk2021ritm,hao2021edgeflow}} &  4.22 & 0 & 8.96 & 0 & 470 & 0  \\
\multicolumn{2}{l|}{hrnet18s-600~\cite{sofiiuk2021ritm,hao2021edgeflow}} &  4.22 & 0 & 20.17 & 0 & 1020 & 0  \\
\multicolumn{2}{l|}{hrnet32-400~\cite{sofiiuk2021ritm}} &  30.95 & 0 & 40.42 & 0 & 1387 & 0  \\
\multicolumn{2}{l|}{resnet50-400~\cite{chen2021cdnet,fbrs,forte2020getting99}} &  31.38 & 0 & 84.63 & 0 & 2359 & 0  \\
\multicolumn{2}{l|}{resnet101-512~\cite{firstclick}} &  50.37 & 0 & 216.55 & 0 & 6267 & 58 \\
\hline
\multicolumn{2}{l|}{Ours-B0-S1} &  3.72 & 0.016 & 0.43 & 0.17 & 41 & 59  \\
\multicolumn{2}{l|}{Ours-B0-S2} &  3.72 & 0.016 & 1.77 & 0.17 & 140 & 60  \\
\multicolumn{2}{l|}{Ours-B3-S2} &  45.6 & 0.025 & 12.72 & 0.20 & 634 & 72 \\
\multicolumn{2}{l|}{Ours-hrnet18s-S1} &  4.22 & 0.011 & 0.91 & 0.15 & 80 & 50  \\
\multicolumn{2}{l|}{Ours-hrnet18s-S2} &  4.22 & 0.011 & 3.66 & 0.16 & 213 & 51  \\
\multicolumn{2}{l|}{Ours-hrnet32-S2} & 30.95  & 0.025 & 16.92 & 0.20 & 650 & 51  \\
\bottomrule[1pt]
\end{tabular}
}
\end{center}
\label{tab:computation}
\vspace{-5mm}
\end{table}

In \cref{tab:computation}, most works use big models and 400 to 600 input sizes, which makes them hard to use on CPU devices. In contrast, FocalClick supports light models and small input sizes like 128 and 256. The FLOPs of our B0-S1 version is 15 times smaller than the lightest RITM~\cite{sofiiuk2021ritm}, 360 times smaller than  FCANet~\cite{firstclick}. Using the same Segmentor, our hrnet-18s version could reduce 2 to 8 times FLOPs compared with original RITM~\cite{sofiiuk2021ritm}. 

Besides, as FocalClick could be adapted to various Segmentors, the FLOPs could be further reduced by using more light-weighted architectures like \cite{howard2017mobilenets,tan2019efficientnet}.\\

\subsection{ Performance for Mask Correction}
We evaluate the performance of mask correction on the DAVIS-585 benchmark, and the results are listed in \cref{tab:interDAVIS}. We also report the results of annotating from scratch. 

All initial masks provided by DAVIS-585 have IOUs between 0.75 to 0.85, and some challenging details have already been well annotated. Hence, making good use of them could logically facilitate the annotation. However, according to  \cref{tab:interDAVIS}, RITM~\cite{sofiiuk2021ritm} do not show much difference between starting from initial masks and scratch.   In contrast, FocalClick makes good use of the initial masks. It requires significantly smaller numbers of clicks when starting from preexisting masks. Besides, it shows that the S1 version of FocalClick could outperform the big version of RITM~\cite{sofiiuk2021ritm} for mask correction tasks with 1/67 FLOPs.

\section{Experiments for \method}
\label{sec:expxl}
In this section, we give a detailed analysis for the extended part of \method. We first report comparison results with state-of-the-art methods across different benchmarks. Afterwards, we report in-depth ablation studies and qualitative demonstrations. 
\begin{table*}[t]
\begin{center}
\caption{ \textbf{Comaprisons with SOTAs.} We report evaluation results on GrabCut, Berkeley, SBD and DAVIS datasets. 
 `NoC~85/90' denotes the average Number of Clicks required the get IoU of 85/90\%. `Synthetic' data uses the datasets of \cite{lin2014coco,xu2017dim,dai2014synthesizability}.  `Large Dataset' denotes a combined dataset.~\cite{lin2014coco,gupta2019lvis,borji2015msra10k,ade20k,wang2017learningduts,xu2018youtubevos,thinobject,cong2020dovenet} }
\vspace{-5pt}
 \label{tab:evaluation sota}
\scalebox{0.98}{
\begin{tabular}{ll|c|c|c|c|c|c|c|c}
\toprule[1pt]
\multicolumn{3}{l|}{} & \multicolumn{2}{c|}{GrabCut~\cite{rother2004grabcut}} & Berkeley~\cite{berkeley} & \multicolumn{2}{c|}{SBD~\cite{SBD}} & \multicolumn{2}{c}{DAVIS~\cite{davis}} \\
\cline{4-10}
\multicolumn{2}{l|}{Method } & Train Data   & NoC~85 & NoC~90 & NoC~90 & NoC~85 & NoC~90 & NoC~85 & NoC~90 \\
\hline
\multicolumn{2}{l|}{Graph cut~\cite{boykov2001interactive}} & /  & 7.98 & 10.00 & 14.22 & 13.6 & 15.96 & 15.13 & 17.41 \\
\multicolumn{2}{l|}{Geodesic matting~\cite{gulshan2010geodesic}} & / & 13.32 & 14.57 & 15.96 & 15.36 & 17.60 & 18.59 & 19.50 \\
\multicolumn{2}{l|}{Random walker~\cite{grady2006random}}  & / & 11.36 & 13.77 & 14.02 & 12.22 & 15.04 & 16.71 & 18.31 \\
\multicolumn{2}{l|}{Euclidean star convexity~\cite{gulshan2010geodesic}}  & /  & 7.24 & 9.20 & 12.11 & 12.21 & 14.86 & 15.41 & 17.70 \\
\multicolumn{2}{l|}{Geodesic star convexity~\cite{gulshan2010geodesic}}  & / & 7.10 & 9.12 & 12.57 & 12.69 & 15.31 & 15.35 & 17.52 \\
\hline
\multicolumn{2}{l|}{DOS w/o GC~\cite{xu2016deep}}  & \scalebox{0.75}{SBD\cite{SBD}}  & 8.02 & 12.59 & -- & 14.30 & 16.79 & 12.52 & 17.11 \\
\multicolumn{2}{l|}{DOS with GC~\cite{xu2016deep}}  & \scalebox{0.75}{SBD\cite{SBD}}& 5.08 & 6.08 & -- & 9.22 & 12.80 & 9.03 & 12.58 \\
\multicolumn{2}{l|}{Latent diversity~\cite{li2018latentdiversity}}  & \scalebox{0.75}{SBD\cite{SBD}} & 3.20 & 4.79 & -- & 7.41 & 10.78 & 5.05  & 9.57 \\
\multicolumn{2}{l|}{RIS-Net~\cite{liew2017regional}}  & \scalebox{0.75}{SBD\cite{SBD}}& -- & 5.00 & -- & 6.03 & -- & -- & -- \\
\multicolumn{2}{l|}{CM guidance~\cite{majumder2019content}}  & \scalebox{0.75}{SBD\cite{SBD}}& -- & 3.58 & 5.60 & -- & -- & -- & --\\
\multicolumn{2}{l|}{BRS~\cite{jang2019brs}}  & \scalebox{0.75}{SBD\cite{SBD}} & 2.60 & 3.60 & 5.08 & 6.59 & 9.78 & 5.58 & 8.24 \\
\multicolumn{2}{l|}{f-BRS-B-resnet50~\cite{fbrs}}  & \scalebox{0.75}{SBD\cite{SBD}} & 2.50 & 2.98 & {4.34} & 5.06 & 8.08 & 5.39 & 7.81 \\
\multicolumn{2}{l|}{ CDNet-resnet50~\cite{chen2021cdnet}}  & \scalebox{0.75}{SBD\cite{SBD}} &  2.22 & 2.64 & 3.69 & 4.37 &  7.87& {5.17} & 6.66 \\
\multicolumn{2}{l|}{ RITM-hrnet18~\cite{sofiiuk2021ritm}}  & \scalebox{0.75}{SBD\cite{SBD}} &  1.76 & 2.04 & 3.22 & \ {3.39} &  \ {5.43} & {4.94} & 6.71 \\
\multicolumn{2}{l|}{SimpleClick-H~\cite{liu2023simpleclick}} & \scalebox{0.75}{SBD\cite{SBD}} & {1.32}   &  {1.44} & {2.09}  & \textbf{2.51}  & \textbf{4.15} & 4.20  & 5.34 \\
\rowcolor{gray!20} 
\multicolumn{2}{l|}{ FocalClick-hrnet18s-S2 } &  \scalebox{0.75}{SBD\cite{SBD}} & 1.86  & 2.06 & 3.14  & 4.30 &  6.52 & \ {4.92} &  {6.48}  \\
\rowcolor{gray!20} 
\multicolumn{2}{l|}{ FocalClick-segformerB0-S2}  & \scalebox{0.75}{SBD\cite{SBD}} & 1.66   &  1.90 & {3.14}  & 4.34  & 6.51 & 5.02  & 7.06 \\
\hline
\multicolumn{2}{l|}{FCANet~(SIS)~\cite{firstclick}}  & \scalebox{0.75}{ SBD\cite{SBD}+PASCAL\cite{everingham2010pascal}} & -    & 2.14 & 4.19 &-     & -    & -    & 7.90 \\
\multicolumn{2}{l|}{ 99\%AccuracyNet~\cite{forte2020getting99}}  & \scalebox{0.75}{SBD\cite{SBD}+Synthetic}   &  - & 1.80 & 3.04 & 3.90 &  - & - & - \\
\multicolumn{2}{l|}{f-BRS-B-hrnet32~\cite{fbrs}}  & \scalebox{0.75}{ COCO\cite{lin2014coco}+LVIS\cite{gupta2019lvis}}  & 1.54 & 1.69 & 2.44 & 4.37 & 7.26 & 5.17 & 6.50 \\
\multicolumn{2}{l|}{ RITM-hrnet18s~\cite{sofiiuk2021ritm}}  & \scalebox{0.75}{ COCO\cite{lin2014coco}+LVIS\cite{gupta2019lvis}} &  1.54 & 1.68 & 2.60 & 4.04 &  6.48 & 4.70 & 5.98 \\
\multicolumn{2}{l|}{ RITM-hrnet32~\cite{sofiiuk2021ritm}}  & \scalebox{0.75}{ COCO\cite{lin2014coco}+LVIS\cite{gupta2019lvis}} &  1.46 & 1.56 & 2.10 & 3.59 & 5.71 & 4.11 & 5.34 \\
\multicolumn{2}{l|}{ EdgeFlow-hrnet18~\cite{hao2021edgeflow}}  & \scalebox{0.75}{ COCO\cite{lin2014coco}+LVIS\cite{gupta2019lvis}} &  1.60 & 1.72 & 2.40 & - &  - & 4.54 & 5.77 \\

\multicolumn{2}{l|}{SimpleClick-B~\cite{liu2023simpleclick}} & \scalebox{0.75}{ COCO\cite{lin2014coco}+LVIS\cite{gupta2019lvis}} & {1.38}   &  {1.48} & {1.97}  & 3.43  & 5.62 & 3.66  & 5.06 \\
\multicolumn{2}{l|}{SimpleClick-H~\cite{liu2023simpleclick}} & \scalebox{0.75}{ COCO\cite{lin2014coco}+LVIS\cite{gupta2019lvis}} & {1.38}   &  {1.50} & {1.75}  &  {2.85}  &  {4.70} & 3.41  & 4.78 \\
\multicolumn{2}{l|}{PseudoClick-HRNet32~\cite{PseudoClick}} & \scalebox{0.75}{COCO\cite{lin2014coco}+LVIS\cite{gupta2019lvis}} & {-}   & {1.50}  & {2.08}  & 4.34  & 5.54 & 3.79  & 5.11 \\
\multicolumn{2}{l|}{InterFormer-Tiny~\cite{huang2023interformer}} & \scalebox{0.75}{ COCO\cite{lin2014coco}+LVIS\cite{gupta2019lvis}} & {-}   &  {1.36} & {2.53}  & 3.25  & 5.51 & -  & 5.21 \\

\rowcolor{gray!20} 
\multicolumn{2}{l|}{ FocalClick-hrnet18s-S1 } &  \scalebox{0.75}{ COCO\cite{lin2014coco}+LVIS\cite{gupta2019lvis}}  &  1.64 & 1.82 & 2.89 & 4.74 & 7.29 & 4.77 & 6.56 \\
\rowcolor{gray!20} 
\multicolumn{2}{l|}{ FocalClick-hrnet18s-S2 } &  \scalebox{0.75}{ COCO\cite{lin2014coco}+LVIS\cite{gupta2019lvis}}   &  1.48 & 1.62 & 2.66 & 4.43 &  6.79 & 3.90 & 5.25 \\
\rowcolor{gray!20} 
\multicolumn{2}{l|}{ FocalClick-hrnet32-S2 } &  \scalebox{0.75}{ COCO\cite{lin2014coco}+LVIS\cite{gupta2019lvis}}   & 1.64  & 1.80  & 2.36 & 4.24 &  6.51 & 4.01 & 5.39 \\
\rowcolor{gray!20} 
\multicolumn{2}{l|}{ FocalClick-segformerB0-S1}  & \scalebox{0.75}{ COCO\cite{lin2014coco}+LVIS\cite{gupta2019lvis}} & 1.60  & 1.86  & 3.29  & 4.98 & 7.60  & 5.13  & 7.42 \\
\rowcolor{gray!20} 
\multicolumn{2}{l|}{ FocalClick-segformerB0-S2}  & \scalebox{0.75}{ COCO\cite{lin2014coco}+LVIS\cite{gupta2019lvis}}  &   {1.40} & 1.66 & 2.27 & 4.56 & 6.86 & 4.04 & 5.49 \\
\rowcolor{gray!20} 
\multicolumn{2}{l|}{ FocalClick-segformerB3-S2}  & \scalebox{0.75}{ COCO\cite{lin2014coco}+LVIS\cite{gupta2019lvis}}  & 1.44  &   {1.50} &  {1.92}  &  {3.53} &  {5.59} &  {3.61} &  {4.90} \\
\hline
\rowcolor{gray!20} 
\multicolumn{2}{l|}{ FocalClick-hrnet32-S2 } &  \scalebox{0.75}{Large Dataset}   & 1.30  & 1.34 & 1.85 & 4.35 &  6.61 & 3.19 & 4.81 \\
\rowcolor{gray!20} 
\multicolumn{2}{l|}{ FocalClick-segformerB3-S2}  & \scalebox{0.75}{Large Dataset} &  1.22 & 1.26 & 1.48 & 3.70 &  5.84 & 2.92 & 4.52 \\
\hline
\multicolumn{2}{l|}{SAM-B~\cite{sam}} & \scalebox{0.75}{SAM Dataset} &  1.82 & 1.86 & 2.17 & 4.97 &  8.20 & 4.20 & 5.12 \\
\multicolumn{2}{l|}{SAM-H~\cite{sam}} & \scalebox{0.75}{SAM Dataset} &  1.72 & 1.80 & 2.09 & 5.23 &  8.50 & 4.24 & 5.34 \\
\multicolumn{2}{l|}{SAM-HQ-H~\cite{hqsam}} & \scalebox{0.75}{SAM Dataset} &  1.80 & 1.88 & 2.17 & 6.30 &  9.41 & 4.23 & 5.12 \\
\rowcolor{gray!20} 
\multicolumn{2}{l|}{FocalClick-XL-B} & \scalebox{0.75}{Combined Dataset} &  1.30 & 1.34 & 1.47 & 3.59 &  5.63  & 2.89 & 4.43\\
\rowcolor{gray!20} 
\multicolumn{2}{l|}{FocalClick-XL-H} & \scalebox{0.75}{Combined Dataset} & \textbf{1.22}  & \textbf{1.24} &  \textbf{1.45} &   {3.32} &   {5.51} &  \textbf{2.66} &  \textbf{4.39} \\
\bottomrule[1pt]
\end{tabular}
}
\end{center}
\vspace{-2mm}
\end{table*}

\subsection{ Quantitative Analysis} 
\noindent \textbf{Click-based segmentation.} We first give a quantitative analysis of \method on click-based interactive segmentation.  In Tab.~\ref{tab:evaluation sota}, we compared \method with previous state-of-the-art methods on GrabCut~\cite{rother2004grabcut}, Berkeley~\cite{berkeley}, SBD~\cite{SBD}, and DAVIS~\cite{davis}.
We divide the table into four blocks according to the usage of training data and observe that our method presents competitive results against the previous state-of-the-art under different settings. 
We could also observe that, compared with the basic FocalClick, our \method demonstrates significant improvements.

In the last block, we list the performance for SAM~\cite{sam} and its variants SAM-HQ~\cite{hqsam}. Although trained on large datasets and with big models, SAM~\cite{sam} does not demonstrate its dominance. As analyzed before, the performance of SAM is limited by fine details. 
SAM-HQ~\cite{hqsam} explores to refine the masks produced by SAM. However, the refinement module is interaction-agnostic. Thus, it could not benefit from the increasing number of user interactions and fails to get satisfactory performance. In contrast, as the magnifier takes the interaction map as guidance, \method could focus on target-centric details and benefits from excessive user interactions.  \method achieves new state-of-the-art across multiple benchmarks.

In \cref{tab:k_iou}, we also report the k-mIoU results on DAVIS~\cite{davis}, HQSeg-44K~\cite{hqsam}, ssTEM~\cite{ssTEM} and BraTS~\cite{brast}. Results demonstrate the strong ability of \method in different evaluation configurations and domains.

\begin{table}[t]
\begin{center}
\caption{ \textbf{Quatitatrive comparisons} for k-mIOU with previous state-of-the-art methods. This metric measures the mean IoU given k numbers of clicks. }
\label{tab:k_iou}
\vspace{-5pt}
\scalebox{0.85}{
\begin{tabular}{l|cccc}
\toprule[1pt]
~ & DAVIS~\cite{davis} & HQSeg-44K~\cite{hqsam} & ssTEM~\cite{ssTEM} & BraTS~\cite{brast} \\
~ & 5-mIoU & 5-mIoU & 10-mIoU & 10-mIoU \\
\hline
RITM~\cite{sofiiuk2021ritm} & 89.75 & 77.72 & \textbf{94.11} & 88.34  \\
FocalClick  & 91.22 & 85.45 & 93.61 & 88.62  \\
SimpleClick~\cite{liu2023simpleclick}  & 90.73 & 85.11 & 93.72 & 86.98  \\
SAM~\cite{sam}  & 90.95 & 86.16 & 91.58 & 87.03  \\
\hline
\rowcolor{gray!20} 
FocalClick-XL  & \textbf{91.89} & \textbf{91.70} & 93.96 & \textbf{88.90}  \\
\bottomrule
\end{tabular}
}
\end{center}
\vspace{-5pt}
\end{table}

\begin{table}[t]
\begin{center}
\caption{ \textbf{Ablation studies for the core components}  on click-based settings. We start with Context-Net~(SAM) as a baseline and add components step by step to verify their effectiveness.}
\label{tab:ablation_all}
\vspace{-8pt}
\scalebox{0.65}{
\begin{tabular}{l|cc|cc|cc}
\toprule[1pt]
\multirow{2}*{Method} & \multicolumn{2}{c|}{GrabCut} & \multicolumn{2}{c|}{Berkeley} & \multicolumn{2}{c}{DAVIS}  \\
~ & NoC~85 & NoC~90& NoC~85 & NoC~90 & NoC~85 & NoC~90 \\
\hline
Context-Net & 1.82 & 1.86 & 1.68 & 2.17 & 4.20 & 5.12 \\
Object-Net  & 1.36 & 1.58 & 1.64 & 2.58 & 4.39  & 5.72 \\
Context-Net + Object-Net   & \textbf{1.28} & 1.52 & 1.48 & 1.92 & 3.64  & 5.01 \\
Context-Net + Object-Net + Detail-Net  & \textbf{1.28} & \textbf{1.48} & \textbf{1.43} & \textbf{1.69} & \textbf{3.29}  & \textbf{4.63} \\
\bottomrule
\end{tabular}
}
\end{center}
\vspace{-15pt}
\end{table}

\begin{table*}[t]
\scriptsize
\begin{minipage}[c]{.48\linewidth}
    \begin{center}
    \caption{ \textbf{Quatitative results~(NoS) of scribble-based interactions.}  ``NoS'' means the number of scribbles required for target IoU. 
    } 
    \vspace{-5pt}
    \label{tab:scribblecompare}
        \scalebox{0.8}{
        \begin{tabular}{l|cc|cc|cc}
        \toprule[1pt]
        \multirow{2}*{Method} & \multicolumn{2}{c|}{GrabCut} & \multicolumn{2}{c|}{Berkeley} & \multicolumn{2}{c}{DAVIS}  \\
        ~ & NoS~85 & NoS~90& NoS~85 & NoS~90 & NoS~85 & NoS~90 \\
        \hline
        AppearanceSim~\cite{appearancesimilarity}   & 5.33 & 6.76 & 5.89 & 6.43 & 8.45  & 14.32 \\
        RITM-Scr-V1    & 2.21 & 3.76 & 2.54 & 3.65 & 7.58  & 8.32 \\
        RITM-Scr-V2    & 1.98 & 2.34 & 1.79 & 2.01 & 5.35  & 6.19 \\
        \hline
        \rowcolor{gray!20} 
        \method-B  & 1.22 & 1.26 & \textbf{1.17} & \textbf{1.35} & 3.03  & 4.51 \\
        \rowcolor{gray!20} 
        \method-H  & \textbf{1.21} & \textbf{1.25} & 1.19 & 1.37 & \textbf{3.01}  & \textbf{4.42} \\
        \bottomrule
        \end{tabular}
        }
    \end{center}
\vspace{-9pt}
\end{minipage}
\hspace{2mm}
\begin{minipage}[c]{.48\linewidth}
    \caption{ \textbf{Quatitative results~(mIoU) of box-based interactions}. 
        We also report the results by providing a single click or scribble as a reference.
        }   
    \vspace{-10pt}
    \label{tab:boxcompare}
    \begin{center}
    \scalebox{0.85}{
        \begin{tabular}{l|c|c|c}
        \toprule[1pt]
        Method  & GrabCut & Berkeley & DAVIS  \\
        \hline
        SAM-B-Click  & 76.16 & 73.00 & 58.75  \\
        SAM-B-Box  & 85.47 & 83.29 & 80.32  \\
        \method-B-Click  & 89.49 & 85.08 & 78.88  \\
        \method-B-Scribble  & 91.93 & 91.24 & 82.66  \\
        \rowcolor{gray!20} 
        \method-B-Box  & \textbf{96.30} & \textbf{94.10} & \textbf{84.89}  \\
        \bottomrule
        \end{tabular}
        }
    \end{center}
\vspace{-0.4cm}
\end{minipage}
\end{table*}

\noindent \textbf{Scribble-based segmentation.}  In \cref{tab:scribblecompare}, we report our performance on scribble-based segmentation.  We use the protocol introduced in \cref{sec:evaluation} to measure the NoS~(Numbers of Scribble). As most of the previous scribble-based methods are not open-sourced, we reproduce some of the representative ones for comparisons. 
In row~1, we develop a similarity-based model like~\cite{appearancesimilarity}, which shows poor performance as it could not deal with the fine details. Row~2 corresponds to using the click-based solution~\cite{sofiiuk2021ritm} to deal with scribbles. We find that, although clicks could be regarded as short scribbles, directly using click-based models could not get satisfactory results. In row~3, we follow IFIS~\cite{IFIS} to simulate scribbles via linking randomly sampled points. This strategy brings improvements compared to using disks but still gets poor performance.
\method shows significant superiority over those naive solutions, which proves the strong transferring abilities of \method across different interaction formats.   We could also refer to Tab.~\ref{tab:evaluation sota} for more analysis.  When both trained on COCO\cite{lin2014coco} and LVIS~\cite{gupta2019lvis}, scribble-based methods are slightly better than clicks-based ones as scribbles could give more indications compared with clicks. 

\noindent \textbf{Boxes and coarse masks as input.}  The box and coarse masks only act as the indication for the initial round. We report the mIoU~(mean Interaction over Union) for the predicted masks.  In \cref{tab:boxcompare}, we report the mean IoU given the single bounding box. \method archives impressive performance with over 90\% mIoU on GrabCut and Berkeley. It shows a notable performance gain against the original version of SAM.
We also report the performance with a single click or scribble. Compared with these two formats, the box gives the best indications for the rough contour of the target object, which would be a premier choice for the first round of interaction. 

In \cref{tab:maskcompare}, we verified the abilities of \method-B to refine the coarse input masks. We conduct different levels of perturbation on the ground truth masks, getting the coarse masks with different IoUs.  The results show that \method is robust for the coarse masks, even poor masks with very low IoUs could be refined to a good state. Following SAM-HQ~\cite{hqsam}, we also make evaluations on BIG~\cite{cheng2020cascadepsp} and UVO~\cite{uvo} in \cref{tab:generalcompare}.

\begin{table*}[t]
\scriptsize
\begin{minipage}[c]{.48\linewidth}
    \begin{center}
    \caption{ \textbf{Quatitative results~(mIoU) of coarse mask refinement.} We give different levels of perturbations on the ground truth masks and report the mIoUs. Results show that \method brings huge improvements to masks with different levels of flaws with robust performance.
    }  
    \vspace{-5pt}
    \label{tab:maskcompare}
        \scalebox{0.85}{
        \begin{tabular}{c|cc|cc|cc}
        \toprule[1pt]
        \multirow{2}*{Pertub-Level} & \multicolumn{2}{c|}{GrabCut} & \multicolumn{2}{c|}{Berkeley} & \multicolumn{2}{c}{DAVIS}  \\
        ~ & Coarse & Refined & Coarse & Refined & Coarse & Refined \\
        \hline
        1 & 88.64 & 98.62 & 86.19 & 96.16 & 85.54  & 92.29 \\
        2 & 78.51 & 97.92 & 77.11 & 95.53 & 76.50  & 91.33 \\
        3 & 69.76 & 97.61 &  67.29 & 94.96 & 67.43  & 90.34  \\
        4 & 58.33 & 96.03 & 59.10 & 93.20  & 58.34  & 88.70 \\
        5 & 49.70 & 89.08 & 42.23 & 81.45  & 41.25 & 78.07  \\
        \bottomrule
        \end{tabular}
        }
    \end{center}
\vspace{-0.5cm}
\end{minipage}
\hspace{3mm}
\begin{minipage}[c]{.45\linewidth}
    \caption{ \textbf{Comparison results with SAM series on out-of-domain datasets}. BIG~\cite{cheng2020cascadepsp} is used to evaluate high-quality segmentation, while UVO~\cite{uvo} covers a large variety of object categories. We use the ground truth box as the prompt in the BIG dataset and use the box prompt on the UVO dataset.
     }   
    \vspace{-10pt}
    \label{tab:generalcompare}
    \begin{center}
        \scalebox{0.99}{
        \begin{tabular}{l|cc|cc}
        \toprule[1pt]
        \multirow{2}*{Method} & \multicolumn{2}{c|}{BIG~\cite{cheng2020cascadepsp}} & \multicolumn{2}{c}{UVO~\cite{uvo}}\\
        ~ & Box & Mask & AP & $AP_B$ \\
        \hline
        SAM~\cite{sam} & 81.1 & 66.6 & 29.7 & 17.3 \\
        SAM-HQ~\cite{hqsam} & 86.0 & 86.9 & 30.1 & 18.5  \\
        \rowcolor{gray!20} 
        \method & 88.3 & 89.1 &  31.2 & 19.3 \\
        \bottomrule
        \end{tabular}
        }
    \end{center}
\vspace{-5pt}
\end{minipage}
\end{table*}

\begin{table*}
\scriptsize
\begin{minipage}[c]{.48\linewidth}
    \begin{center}
    \caption{ \textbf{Ablation studies for adapting the Context-Net.} We try placing the adapters at different positions and pick the one with the highest performance. 
    } 
    \vspace{-5pt}
    \label{tab:samadapter}
        \scalebox{0.8}{
        \begin{tabular}{l|cc|cc|cc}
        \toprule[1pt]
        \multirow{2}*{Adapter} & \multicolumn{2}{c|}{GrabCut} & \multicolumn{2}{c|}{Berkeley} & \multicolumn{2}{c}{DAVIS}  \\
        ~ & NoC~85 & NoC~90& NoC~85 & NoC~90 & NoC~85 & NoC~90 \\
        \hline
        None & 1.36 & 1.66 & 1.62 & 2.01 & 4.36  & 5.33 \\
        Last layer  & 1.38 & 1.72 & 1.52 & 1.88 & 4.30  & 5.20  \\
        Each stage   & \textbf{1.28} & \textbf{1.48} & \textbf{1.43} & \textbf{1.69} & \textbf{3.29}  & \textbf{4.63} \\
        \bottomrule
        \end{tabular}
        }
    \end{center}
\vspace{-0.4cm}
\end{minipage}
\hspace{0mm}
\begin{minipage}[c]{.48\linewidth}
    \caption{ \textbf{Ablation studies for the transfer tuning strategies}. From the click-based setting to the scribble-based setting, we explore tuning different parts of the model.
    } 
    \vspace{-10pt}
    \label{tab:tuning}
    \begin{center}
        \scalebox{0.75}{
        \begin{tabular}{l|cc|cc|cc}
        \toprule[1pt]
        \multirow{2}*{Method} & \multicolumn{2}{c|}{Berkeley} & \multicolumn{2}{c|}{DAVIS} & \multicolumn{2}{c}{ADE-full}  \\
        ~ & NoC~85 & NoC~90& NoC~85 & NoC~90 & NoC~85 & NoC~90 \\
        \hline
        Full Object-Net  & \textbf{1.22} & \textbf{1.26} & \textbf{1.14} & \textbf{1.28} & 3.04  & 4.55 \\
        Object-Net Decoder  & 1.26 & 1.33 & 1.23 & 1.46 & 3.45  & 4.71 \\
        Prompting Layer   & \textbf{1.22} & \textbf{1.26}  & 1.17 & 1.35 & \textbf{3.03}  & \textbf{4.51}  \\
        \bottomrule
        \end{tabular}
        }
    \end{center}
\vspace{-5pt}
\end{minipage}
\end{table*}

\begin{table*}[t]
\scriptsize
\begin{minipage}[c]{.49\linewidth}
    \begin{center}
    \caption{ \textbf{Computation analysis}. Our methods introduce a modest increase in parameters. However, the performance gains are significant.
    }   
    \vspace{-5pt}
    \label{tab:params}
        \scalebox{0.98}{
        \begin{tabular}{l|c|c|c}
        \toprule[1pt]
        Method  & Params  & NoC90~(SBD) & NoC90~(DAVIS)   \\
        \hline
        SAM-B    &  94M   & 8.20 & 5.12  \\
        SAM-H    & 641M   & 8.50 & 5.34  \\
        \rowcolor{gray!20} 
        FocalClick-XL-B  &  145M  & 5.63 & 4.43 \\
        \rowcolor{gray!20} 
        FocalClick-XL-H  &  693M  & \textbf{5.51} &  \textbf{4.39}\\
        \bottomrule
        \end{tabular}
        }
    \end{center}
\vspace{-0.4cm}
\end{minipage}
\hspace{2mm}
\begin{minipage}[c]{.49\linewidth}
    \caption{ \textbf{Speed analysis}. The online inference speed~(waiting time after each interaction) is comparable to or even faster than previous works.
    }   
    \vspace{-10pt}
    \label{tab:speed}
    \begin{center}
        \scalebox{0.98}{
        \begin{tabular}{l|c|c}
        \toprule[1pt]
        Method  & Speed~(Online)   & NoC90~(DAVIS)   \\
        \hline
        f-BRS~\cite{fbrs}    &  320~ms  & 7.81  \\
        CDNet~\cite{chen2021cdnet}    & 230~ms  & 6.66  \\
        SimpleClick~\cite{liu2023simpleclick}    & 132~ms   & 4.78  \\
        \rowcolor{gray!20} 
        FocalClick-XL-B  &  220~ms  & 4.43 \\
        \bottomrule
        \end{tabular}
        }
    \end{center}
\vspace{-5pt}
\end{minipage}
\end{table*}

\begin{figure*}[t]
\centering 
\includegraphics[width=1.0\linewidth]{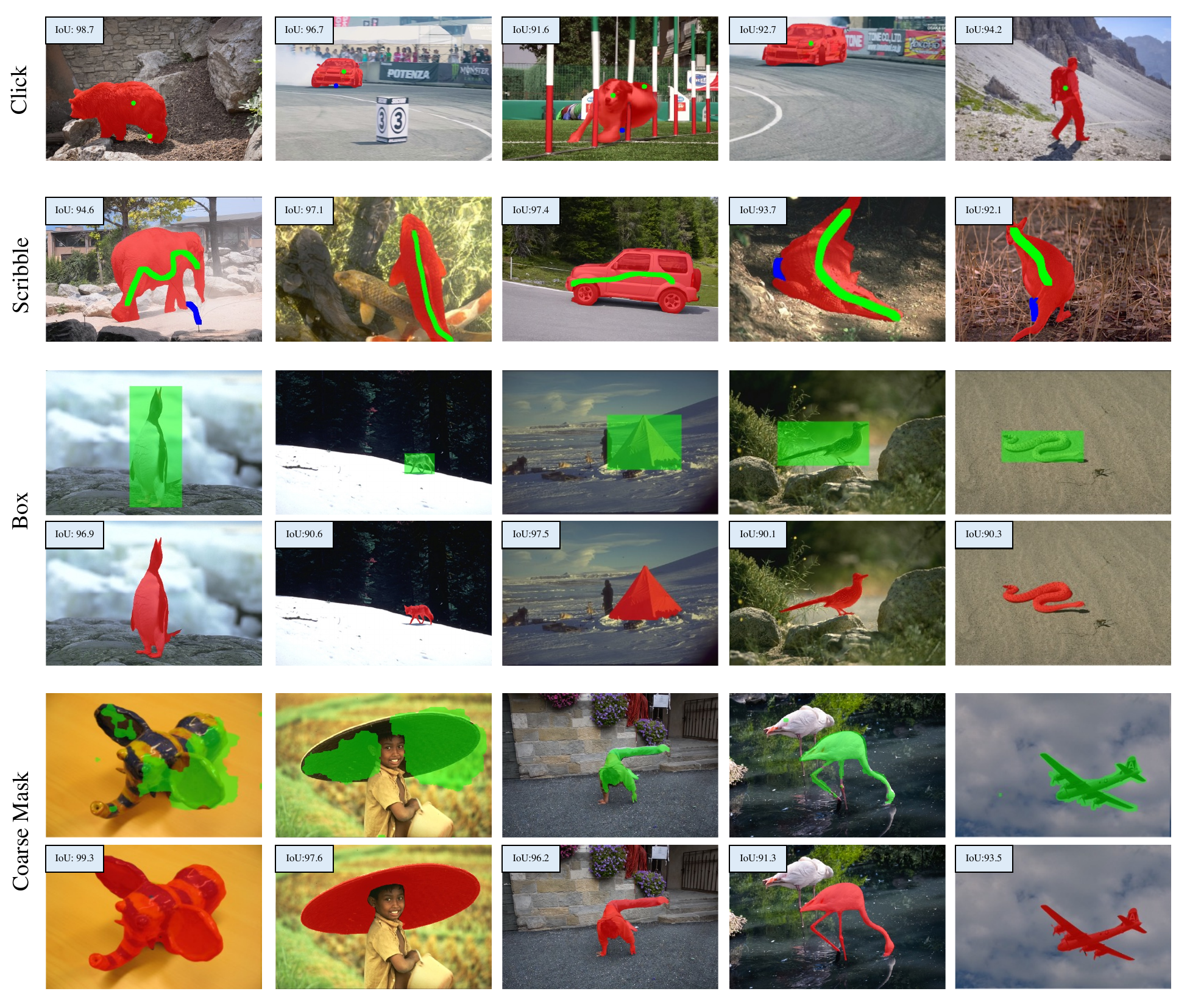} 
\vspace{-23pt}
\caption{%
    \textbf{Demonstrations for \method.} Our method provides a unified solution for various interaction formats and predicts high-quality masks with fine details.
}
\label{fig:demo}
\vspace{-10pt}
\end{figure*}

\begin{figure}[t]
\centering
\includegraphics[width=1.0\linewidth]{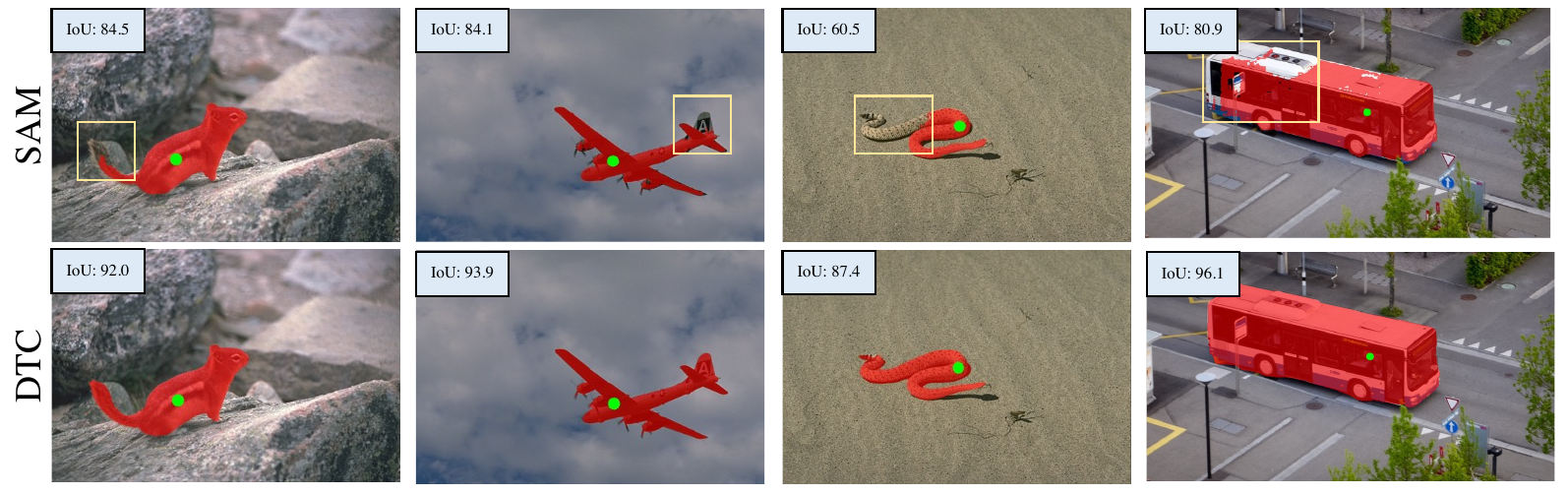} 
\vspace{-18pt}
\caption{%
    \textbf{Qualitative comparisons with SAM.} \method shows significantly better mask qualities compared with SAM given a single click.
}
\label{fig:compsam}
\vspace{-2mm}
\end{figure}

\subsection{Ablations Studies}
After verifying our promising performance, in this section, we dive into the details of our framework. We conduct ablation studies to verify the effectiveness of our design.
We first analyze the importance of each subnet of our pipeline. Afterward, we make experiments for the structure of the SAM adapter and the tuning strategy to transfer \method from click to other interactions.

\noindent \textbf{Task decomposation.}
We analyze each subnet of \method in \cref{tab:ablation_all}. The Context-Net and Object-Net could both be utilized as a baseline to achieve the function of interactive segmentation individually. We first report their performance in the first two columns. Afterward, we add each subnet step by step.
The results demonstrate steady improvements brought by each component.

\noindent \textbf{SAM adapter.}
We find that the adapter layer in the Context-Net plays a crucial role in leveraging the strong context-aware priors of SAM for our task. In \cref{tab:samadapter}, we explore several design choices. The results indicate that incorporating additional Conv-GeLU-Conv layers after each transformer block yields the most favorable performance.

\noindent \textbf{Transfer tuning.} 
We further investigate how to adapt a click-based \method to other forms of user interactions, using scribbles as an example. In \cref{tab:tuning}, we explore tuning different components of the model. Our findings reveal that adjusting just a single input layer is sufficient to transfer \method across tasks. This suggests that most of the knowledge is shared across different interaction types, and the key lies in effectively encoding these interactions into the corresponding control signals.

\noindent \textbf{Computation analysis.} In \cref{tab:params}, we report the numbers of parameters of \method and SAM. Our method requires acceptable additional parameters but brings obvious performance improvement. In \cref{tab:speed}, we compare the inference speed with previous state-of-the-art interactive segmentation models. Thanks to the progressive magnification strategy, our \method shows comparable or better speed for the time of waiting after each interaction.

\subsection{Qualitative Analysis}
We begin by visualizing the comparison results with SAM in \cref{fig:compsam}, where our method demonstrates significant improvements. Leveraging Object-Net and Detail-Net, \method achieves high-quality segmentation with minimal user interactions.

In \cref{fig:demo}, we highlight the generalization capabilities of \method across different interaction types. The model accurately refines masks with iteratively added clicks and scribbles (rows 1–2). While box-based interactions typically support single-round refinement, they effectively capture the target's overall location and shape, leading to satisfactory segmentation results. Additionally, \method exhibits strong robustness when processing coarse masks—regardless of the input mask quality, it consistently enhances segmentation accuracy.

\begin{figure*}[!t]
\vspace{10pt}
\centering 
\includegraphics[width=1.0\linewidth]{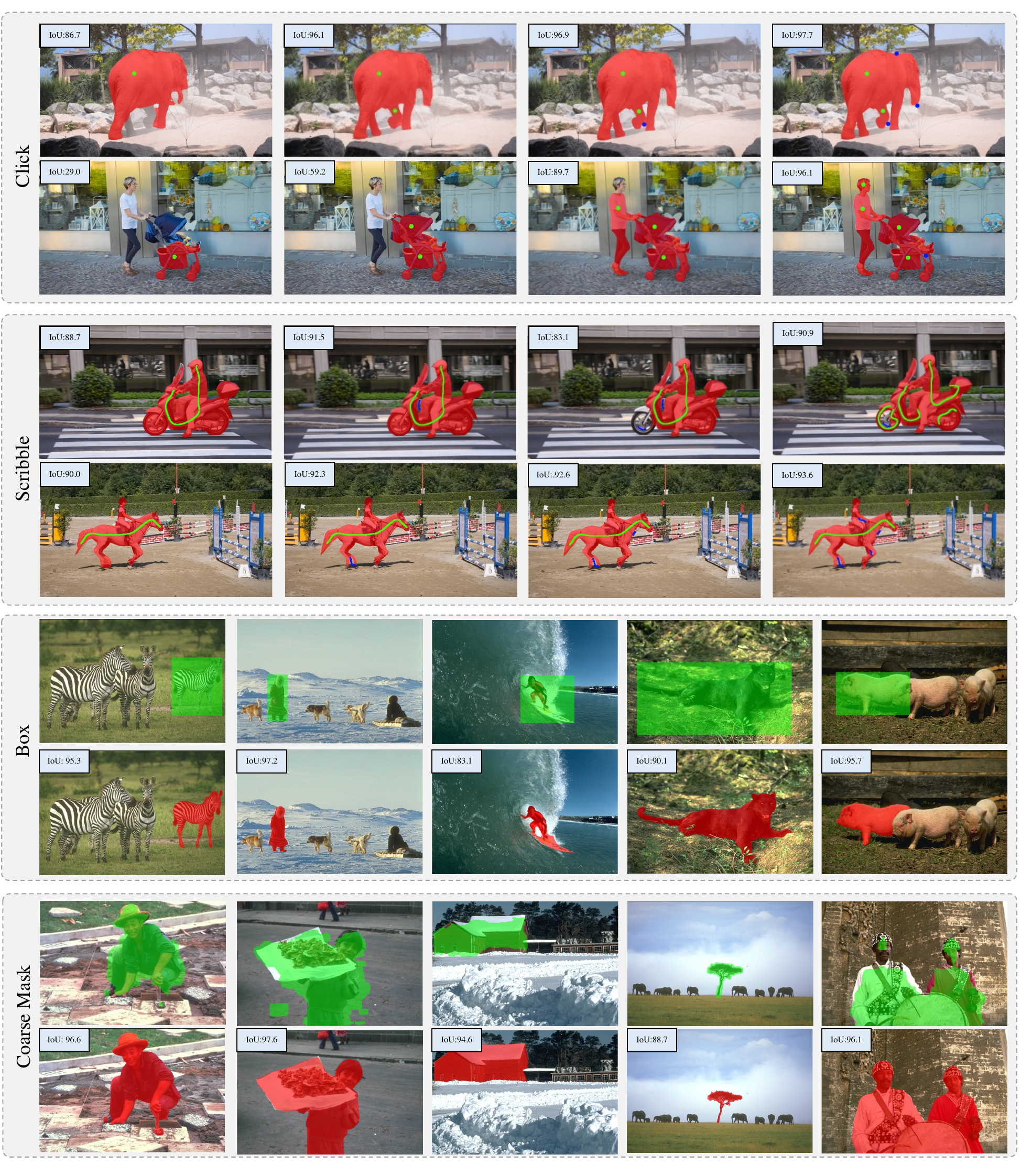} 
\vspace{-20pt}
\caption{ 
    \textbf{More visualization results for \method.} For the clicks and scribbles, the interactions are iteratively added at the center of the maximum error region. For the boxes and coarse masks, the upper row shows the user interactions, and the bottom row denotes the prediction results.  
}
\label{fig:demo_more}
\vspace{5pt}
\end{figure*}

Further prediction results are shown in \cref{fig:demo_more}, providing additional evidence of \method's effectiveness in handling diverse user interactions. Clicks and scribbles are sequentially integrated to iteratively refine earlier predictions, demonstrating the model’s adaptability to various user input formats.

\section{Conclusion}
In this paper, we build upon the classical interactive segmentation framework, FocalClick, and introduce significant extensions.
We scale up its core pipeline and expand its capabilities to support a broader range of user interactions, presenting \method. This enhanced pipeline decomposes interactive segmentation into distinct subtasks—context-level, object-level, and detail-level modeling. Such a structured decomposition allows each component to be thoroughly pre-trained, unlocking the full potential of the model. Moreover, we introduce interaction-related guidance only at the object level, enabling a seamless transition from click-based segmentation to other interaction formats such as boxes, scribbles, and coarse masks.
Beyond architectural advancements, we also examine the evaluation protocols for different interaction types, proposing new benchmarks and metrics to standardize performance assessment.

\noindent \textbf{Limitations and Future Directions.}
While \method demonstrates strong generalization across various interaction types, the current version does not yet support language-based descriptions or example images as prompts. Extending \method to incorporate these modalities will be a key direction for future research, further enhancing its versatility as a unified and flexible segmentation tool.
\\
{\small
\bibliographystyle{unsrt}
\bibliography{reference.bib}
}
\end{document}